\documentclass[3p,times]{elsarticle}
\usepackage{lineno,hyperref}
\usepackage{amsmath,amssymb,amsfonts}
\usepackage{algorithmic}
\usepackage{graphicx}
\usepackage{caption}
\usepackage{textcomp}
\usepackage{multirow}
\usepackage{enumitem}
\usepackage{array}
\usepackage{longtable}
\usepackage{natbib}
\usepackage{lscape}
\usepackage{multicol}
\setcitestyle{square, comma, round, sort&compress, numbers}
\usepackage{geometry}
\biboptions{authoryear}
\begin{document}

\begin{frontmatter}

\title{Automatic Speech Recognition And Limited Vocabulary: A Survey \tnoteref{mytitlenote}}
%\tnotetext[mytitlenote]{Fully documented templates are available in the elsarticle package on \href{http://www.ctan.org/tex-archive/macros/latex/contrib/elsarticle}{CTAN}.}

%
\author[1]{Jean Louis K. E. Fendji\corref{mycorrespondingauthor}} 
\cortext[mycorrespondingauthor]{Corresponding author}
\ead{lfendji@gmail.com}
\author[2]{Diane M. Tala} 
\author[1]{Blaise O. Yenke}
\author[3]{Marcellin Atemkeng}
\address[1]{Department of Computer Engineering, University Institute of Technology, University of Ngaoundere, 455 Ngaoundere, Cameroon (e-mail: lfendji@gmail.com, byenke@yahoo.com)}
\address[2]{Department Mathematics and Computer Science, Faculty of Science, University of Ngaoundere, 454 Ngaoundere, Cameroon (e-mail: talametalomdiane@gmail.com)}
%\fntext[myfootnote]{Since 1880.}
\address[3]{Department of Mathematics, Rhodes University, Grahamstown 6140, South Africa (e-mail: m.atemkeng@ru.ac.za)}

%% or include affiliations in footnotes:  \fnref{myfootnote}

%\ead[url]{www.elsevier.com}

%\author[mysecondaryaddress]{Global Customer Service\corref{mycorrespondingauthor}}
%\cortext[mycorrespondingauthor]{Corresponding author}
%\ead{support@elsevier.com}

%\address[mymainaddress]{1600 John F Kennedy Boulevard, Philadelphia}
%\address[mysecondaryaddress]{360 Park Avenue South, New York}

\begin{abstract}
Automatic Speech Recognition (ASR) is an active field of research due to its large number of applications and the proliferation of interfaces or computing devices that can support speech processing. However, the bulk of applications are based on well-resourced languages that overshadow under-resourced ones. Yet, ASR represents an undeniable means to promote such languages, especially when designing human-to-human or human-to-machine systems involving illiterate people. An approach to design an ASR system targeting under-resourced languages is to start with a limited vocabulary. ASR using a limited vocabulary is a subset of the speech recognition problem that focuses on the recognition of a small number of words or sentences. This paper aims to provide a comprehensive view of mechanisms behind ASR systems as well as techniques, tools, projects, recent contributions, and possible future directions in ASR using a limited vocabulary. This work consequently provides a way forward when designing an ASR system using limited vocabulary. Although an emphasis is put on limited vocabulary, most of the tools and techniques reported in this survey can be applied to ASR systems in general.\end{abstract}

\begin{keyword}
\texttt{Deep learning, Dataset, Machine Learning, Limited Vocabulary, Speech Recognition}

\end{keyword}

\end{frontmatter}

%\linenumbers

%\titlepgskip=-15pt

%\maketitle

\begin{multicols}{2}
\section{Introduction}
\label{sec:introduction}

Automatic speech recognition (ASR) is the process and the related technology applied to convert a speech signal into the matching sequence of words or other linguistic entities using algorithms implemented in computing devices  \citep{b1} . ASR has become an exciting field for many researchers. Presently, users prefer to use devices such as computers, smartphones, or any other connected device through speech. 
Automatic speech recognition can technically be defined as a power density spread over a time-frequency domain \citep{b2}. Current speech processing techniques (encompassing speech synthesis, speech processing, speaker identification or verification) pave the way to create human-to-machine voice interfaces. ASR can be applied in several applications including  voice services \citep{b3}, program control and data entry \citep{b3}, avionics \citep{b3}, disabled assistance \citep{b6,b7}, amongst others.
Although ASR can be advantageous in easing human-to-machine communication; in many cases, it is goes beyond helpful and becomes absolutely necessary. For example, low-literacy levels and the extinction of under-resourced languages are ideal candidates for ASR.\citep{b8}. In fact, the high penetration of communication tools such as smartphones in the developing world \citep{b9} and their increasing presence in rural areas \citep{b10,b11} provides an unprecedented opportunity to develop a voice-based application that can help to mitigate the low literacy levels in those areas. Smartphones offer many advantages over a PC-based interface, such as high mobility and portability, easy recharge of their batteries, and conventional embedded features such as microphones and speakers.

\subsection{Motivation}

In regions with low literacy levels, people are used to speaking local languages that are often considered as under-resourced languages because of the lack or insufficiency of formal written grammar and vocabulary. Since people do not know how to read or to write well-resourced languages (such as English or French), the development of ASR systems for under-resourced languages appears as an appealing solution to overcome this limitation. However, due to the complexity of the task, limited vocabulary must be considered. 

This paper focuses on limited vocabulary in ASR to allow researchers who wish to work on under-resourced languages to have an overview on how to develop a speech recognition system for limited vocabulary. In contrast to limited vocabulary systems, large vocabulary continuous speech recognition (LVCSR) systems are usually trained on thousands of hours of speech and billions of words of text \citep{b12}. The development of large vocabulary systems is complex since the larger the vocabulary, the harder the manipulation of learning algorithms, with more rules needed to build the dataset. LVCSR systems can be very efficient when they are applied on similar domains to those on which they were trained \citep{b13}. However, they are not robust enough to handle mismatched training and test conditions as the context may not be well handled. In fact, most of the input can be silence or contain background noise, which can be mistaken for speech; this increases the false positive rate \citep{b14}. Thus, LVCSR systems are not suitable for transfer learning targeting small or limited vocabulary.

\subsection{Position with other surveys}
Extensive research has been done regarding speech recognition using limited vocabulary. Among the existing surveys, authors in \citep{b15} focus on Portuguese-based language (and variations). The authors consider Portuguese as an understudied language compared to English, Arabic, and Asian languages. Among Asian languages, Indian languages received particular attention. Works on the development of ASR systems dealing with Indian languages such as Hindi, Punjabi, Tamil, amongst other, are presented in \citep{b16}. Another specific language survey is provided in \citep{b17}, where the authors focus on Russian language specificities and apply models for the development of Russian speech recognition systems in some organisations, both in Russia and abroad. A broader survey inspecting more than 120 promising works on biometric recognition (including voice) based on deep learning (DL) models is provided in \citep{b18}. In the latter, the authors present the strengths and possible uses of DL. A narrowed survey is presented in \citep{b19,b163} and highlights the major subjects and improvements made in ASR. Additionally, a conical survey focusing on various feature extraction techniques in speech processing is provided in \citep{b20}. The work in \citep{b21} attempts to provide a comprehensive survey on noise-resistant features as well as similarity measurement, speech enhancement, and speech model compensation in noisy contexts. Due to the increasing penetration rate of mobile devices, the work in \citep{b22} investigates different approaches for providing ASR technology to mobile users. Approaches used to design Chatbots and a comparison between different design techniques from nine papers are given in \citep{b23}. To evaluate the reliability of recognition results, the work in \citep{b24} summarises most research works related to confidence measures. One of the first comprehensive surveys on speech recognition systems for under-resourced languages is found in \citep{b8}, notwithstanding the fact that many of the issues and approaches presented in the  paper apply to speech technology in general. In this survey, authors do not focus on limited vocabulary. The authors in \citep{b162} have done a review for ASR for small vocabulary, but in this survey they do not clearly describe the methods and techniques used to build the models. Moreover this survey is limited to the Marathi language only. Table \ref{tab1} provides a summary of the recent surveys on ASR.

\begin{table*}
\caption{Summary of the recent surveys on ASR with objective and limitation.}\label{tab1}
\centering
\begin{tabular}{ p{2cm} p{4cm} p{4cm} p{2.5cm} p{2.5cm}}
 \hline
 Reference Year	& Objective of the review &	Limitation&	Language specific& 	Limited vocabulary\\
 \hline
 \citep{b21} &
To present noise resistant features and similarity measurement, speech enhancement and speech model compensation in noisy environments &	No  description of the construction mechanism of an ASR.	&No	&No\\

\citep{b17} &
To expose specificities, methods, and applied models for the development of Russian speech recognition.&	Although the authors provided a broad view of works related to ASR for Russian, they did not describe the ASR development methods. They focused on systems and their performance.&	Russian &	No\\

\citep{b22} &
To detail approaches for providing ASR technology to mobile users &	Only interested in ASR architecture. No insight on dataset or corpora. &No	&No \\

\citep{b19} &
To summarise some of the well-known methods used in several stages of ASR systems.&	Provide insight on speech recognition techniques, feature extraction techniques and traditional performance metrics.&	No	&No\\

\citep{b16}&
To present major research works in development of ASR in Indian languages. &	Focuses on Indian languages and does not describe either ASR development methods or dataset construction.&	Indian &	No\\
 
\citep{b20}&
To present various feature extraction techniques in speech processing.&	Focuses solely on extraction techniques.&	No&	No\\

\citep{b8}&
To discuss speech recognition system for under-resourced languages.&	Provides a good understanding of under-resourced languages and mechanism for ASR, but does not describe dataset design mechanism.&	No	&No\\

\citep{b23}&
To present techniques used to design chatbots.&	Does not detail the speech recognition mechanism or the procedure for creating dataset.&	No &	No\\

\citep{b15}&
To present Portuguese-based ASR.&	Focuses only on Portuguese-based ASR (corpora, approaches), even if tools can be used regardless of the language.&	Portuguese	&No\\

\citep{b18}&
To provide an insight of Biometric recognition including voice and based on DL. models	Not specific to speech recognition.& Only provides insight with no description of the ASR mechanism or dataset construction.&	No&	No\\

\citep{b162}&
To review ASR system or interfaces for specific tasks.& Limited to ASR for small vocabulary for Marathi language only.&	Marathi &	Yes\\

\citep{b163}&
To present a comparison between ASR techniques nowadays used and deep learning methods .& The authors do not dwell on the dataset construction techniques used in the different papers cited.&	No &	No\\
 \hline
\end{tabular}

\end{table*}

\subsection{Contributions}
\begin{figure*}
\centering
\includegraphics[width=14cm]{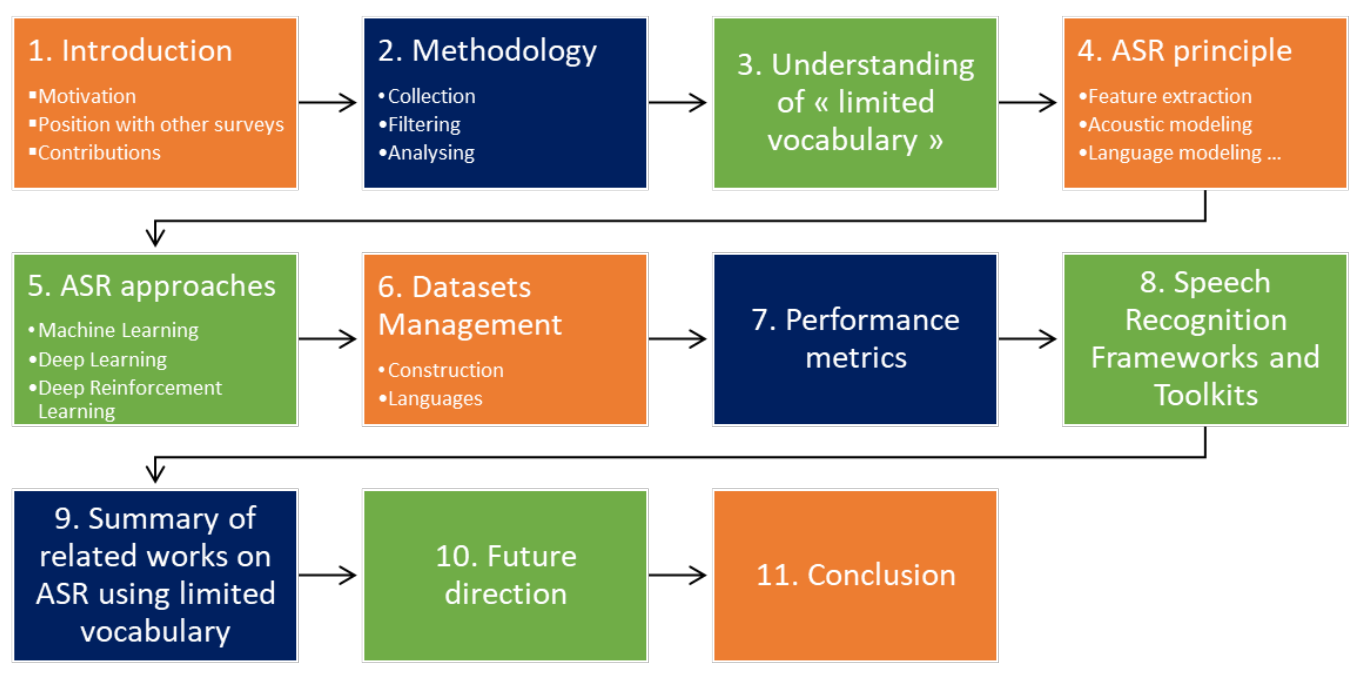}
\caption{Organisation of the paper.}\label{fig1}
\end{figure*}

Despite the plethora of surveys, a survey on ASR using limited vocabularies is yet to be conducted. However, ASR using limited vocabulary is a tremendous opportunity as a starting point for the development of speech recognition systems for under-resourced languages. This survey helps to fill this gap by presenting a summary of works done on ASR for limited vocabulary. For a better understanding, the ASR principle is detailed along with the approach to i) build ASR systems, ii) construct datasets and iii) evaluate the performance of such systems. Furthermore, close and open-source toolkits, and frameworks are also presented. Therefore, such a study can rapidly and easily enable researchers who want to build speech recognition systems using limited vocabulary. The contributions of this paper are as follows:
\begin{itemize}
    \item A description of fundamental aspects of ASR;
     \item A description of tools and processes for creating ASR systems;
     \item A summary of important contributions in ASR with limited vocabulary; and
     \item Orientations for future works.
\end{itemize}
The rest of the paper is organised around eleven sections as shown in Figure~\ref{fig1}. Section 2 presents the methodology used to conduct this survey. Section 3 provides an understanding of “limited vocabulary”, and Section 4 describes the principle of speech recognition. Section 5 describes the techniques used for ASR, and Section 6 deals with the management of datasets. Section 7 presents the traditional performance metrics, and Section 8 provides an insight into the speech recognition frameworks. Section 9 summarises works on speech recognition using limited vocabulary. Section 10 discusses possible future directions, followed by a conclusion in Section 11. The list of abbreviations used throughout this work is shown in Table~\ref{tab2}.

\begin{table*}
\caption{List of Abbreviations}\label{tab2}
\centering
\begin{tabular}{ p{1.5cm} p{5.5cm} |p{1.8cm} p{6.3cm}}
 \hline
 Nomenclature & Definition&  Nomenclature &Definition\\
 \hline
ACC: &Accuracy &AM: &Acoustic Model\\

ASR: &Automatic Speech Recognition& BD-4SK-ASR:& Basic Dataset for Sorani Kurdish Automatic Speech Recognition\\

CER:& Character Error Rate&
CMU:& Carnegie Mellon University\\

CNN:& Convolutional Neural Network&
CNTK:& CogNitive ToolKit\\

CUED:& Cambridge University Engineering Department&
DCT:& Discrete Cosine Transformation\\

DL:& Deep Learning&
DNN:& Deep Neural Network\\

DRL:& Deep Reinforcement Learning&
DWT:& Discrete Wavelet Transform\\

FFT:& Fast Fourier Transformation&
GMM:& Gaussian Mixture Model\\

HMM:& Hidden Markov Model&
HTK:& Hidden Markov Model ToolKit\\

JASPER:& Just Another Speech Recognizer&
LDA:& Linear Discriminant Analysis\\

LER:& Letter Error Rate&
LGB:& Light Gradient Boosting Machine\\

LM:& Language Model&
LPC:& Linear Predictive Coding\\

LVCSR:& Large Vocabulary Continuous Speech Recognition&
LVQ: &Learning Vector Quantization Algorithm\\

MFCC: & Mel-Frequency Cepstrum Coefficient&
ML:& Machine Learning\\

PCM:& Pulse-Code Modulation&
PPVT:& Peabody Picture Vocabulary Test\\

RASTA:& RelAtive SpecTral&

RLAT:& Rapid Language Adaptation Toolkit\\
S2ST: & Speech-to-Speech Translation&

SAPI:& Speech Application Programming Interface\\
SDK:& Software Development Kit&

SVASR: &Small Vocabulary Automatic Speech Recognition\\
WER:& Word Error Rate\\

\hline
\end{tabular}

\end{table*}

\section{Methodology}
The research methodology is composed of three main steps, namely the collection of papers, filtering to keep only relevant papers with significant findings, and analysis of the selected papers. The procedure used for collection and filtering are detailed below.
\subsection{Collection}
The collection of papers has been performed by a keyword-based search from common sources, such as Scopus, IEEE Xplore, ACM, ScienceDirect, PubMed. In addition, the search has been extended to web scientific indexing services, namely Web of Science and Google Scholar.  The aim was to collect as many relevant papers as possible regarding ASR.  For that purpose, several keywords have been used, such as:  “Speech Recognition”] AND [“limited vocabulary” OR “vocabulary” OR “commands”].

\subsection{Filtering}
\begin{figure*}
\centering
\includegraphics[width=14cm]{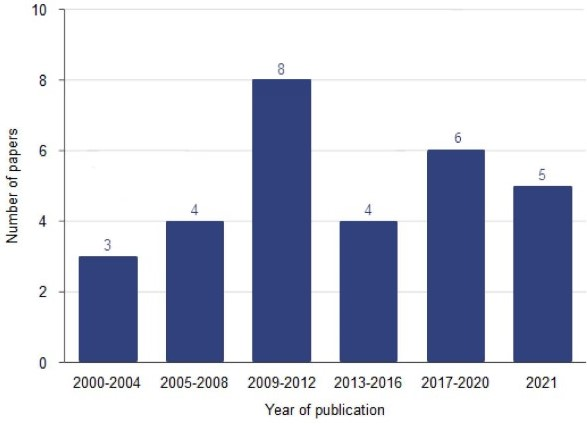}
\caption{Number of relevant papers from speech recognition using limited vocabulary analysed in this survey with respect to publication year.}\label{fig2}
\end{figure*}

This step consists of identifying relevant papers by reading the abstract and screening the papers. Papers that do not directly deal with the topic or provide substantial contributions have been removed. 
However, some of papers providing relevant results have not been considered for the synthesis because they did not provide most of the needed parameters, such as the speech recognition language, the error rate, the type of environment, the exact size of the vocabulary and the number of speakers. We have selected 30 papers for data synthesis. Figure~\ref{fig2} presents the number of articles selected that have been published between 2000 and 2021. 
\subsection{Analysing}
Each paper has been analysed based on the following criteria:
\begin{itemize}
    \item The language for which the recognition system was designed;
    \item The toolkit used to design the models;
    \item The size of the dataset;
    \item The type of environment (noisy or not) for the recognition;
    \item Number of speakers considered for the elaboration of the dataset; and
    \item The accuracy or recognition rate.
\end{itemize}

\section{Understanding of  Limited vocabulary }
A vocabulary, in our context, is defined as a closed list of lexical units that can be recognised by an ASR system. The size of the vocabulary and the selection of lexical units in the vocabulary strongly influence the performance of the automatic transcription system since not all words outside the vocabulary can be recognised by the system.  Speech recognition systems basically access a dictionary of phonemes and words. Obviously, it is easier to seek the meaning of one out of ten words in a ten-word dictionary rather than one out of thousands of words in a Webster's dictionary \citep{b25}. 

A phoneme is, in essence, the smallest unit of phonetic speech that distinguishes one word from another \citep{b26}. Every word can be de-constructed into units (phonemes) of individual sounds that constitutes that word. The number of words or sentences an ASR system can recognise is an important classification criterion. As proposed in \citep{b27}, an ASR system can be classified based on the size of the vocabulary as follows: 
\begin{itemize}                                                                     
    \item 	Small Vocabulary: between 1 and 100 words or sentences;
    \item Medium Vocabulary: between 101 and 1000 words or sentences;
    \item Large Vocabulary: between 1001 and 10,000 words or sentences; and
    \item Very-large vocabulary: more than 10,000 words or sentences.
\end{itemize}

A second classification is proposed by Whittaker and Woodland \citep{b28}. They also classified ASR systems into four categories: small vocabulary (up to 1000 words), medium vocabulary (up to 10,000 words), large vocabulary (up to 100,000 words) and very/extra large vocabulary (up to 100,000 words). In this second case, limited vocabulary can be considered as a small vocabulary meaning up to 1000 words. This size will be used in the rest of the paper as the maximum size for limited vocabulary.

\section{Automatic speech recognition }
In the basic principle of ASR, the person speaking emits pressure variations in his larynx. The sounds produced are digitised by the microphone and transmitted through a medium or a network. Digitised sounds are transformed into acoustic units (or acoustic vectors) via an acoustic model (AM). Thereafter, the recognition engine analyses this sequence of acoustic vectors by comparing it with those in its memory (its language model [LM]) and proposes the most likely candidate sequence. It is therefore necessary for the sequence of acoustic vectors to approximate one of the sequences memorised by the recognition engine. The core of an ASR system can be seen as a mathematical model that can generate a text corresponding to the recognised pieces of speech input \citep{b29}. 

\subsection{Architecture}
Audio signals need to be digitised before the recognition process starts. The digitisation of the signal requires the selection of an appropriate sampling frequency to catch the high-pitched voices \citep{b30}. In general, all ASR systems have the same architecture whether the vocabulary is limited, medium, large, or very large. This architecture can be modified or supplemented according to the recognition to be performed. ASR is usually composed of five typical components:
\begin{enumerate}
    \item Feature extraction;
    \item Acoustic Model;
    \item Language Model;
    \item Pronunciation Model; and
    \item Decoder.
\end{enumerate}

\begin{figure*}
\centering
\includegraphics[width=16cm]{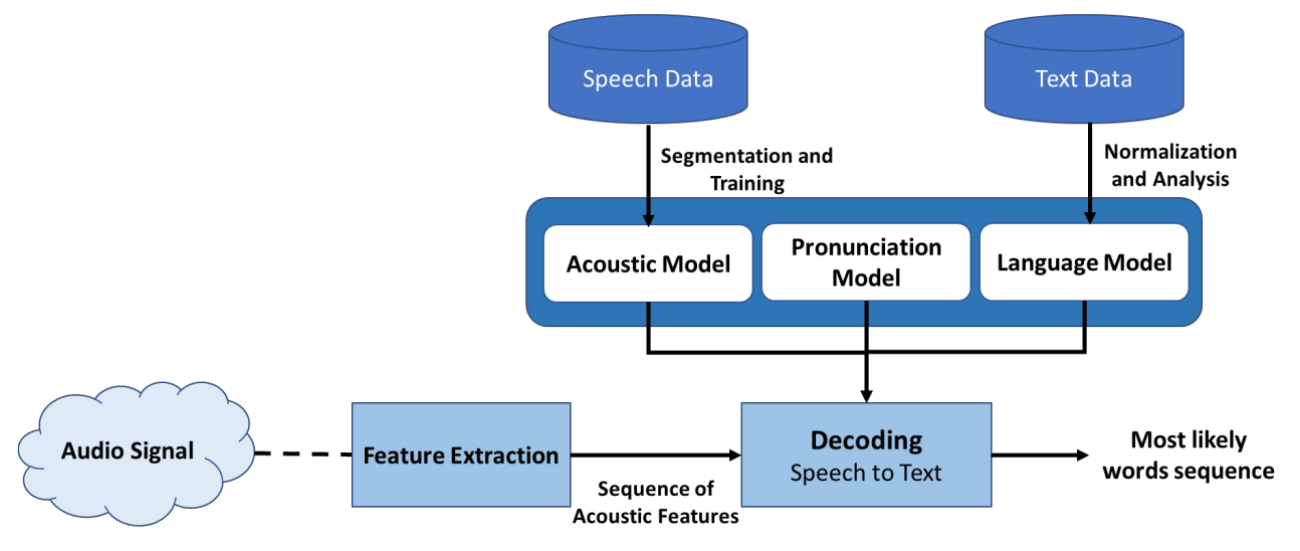}
\caption{Architecture of an ASR (adapted from \cite{b163}.}\label{fig3}
\end{figure*}

In the architecture illustrated in Figure~\ref{fig3}, the speech signal is received and then features are extracted. The obtained parameters are passed to the decoder, which uses the language, the acoustic, and the pronunciation models for learning. 

\subsection{Feature extraction}
\begin{figure*}
\centering
\includegraphics[width=16cm]{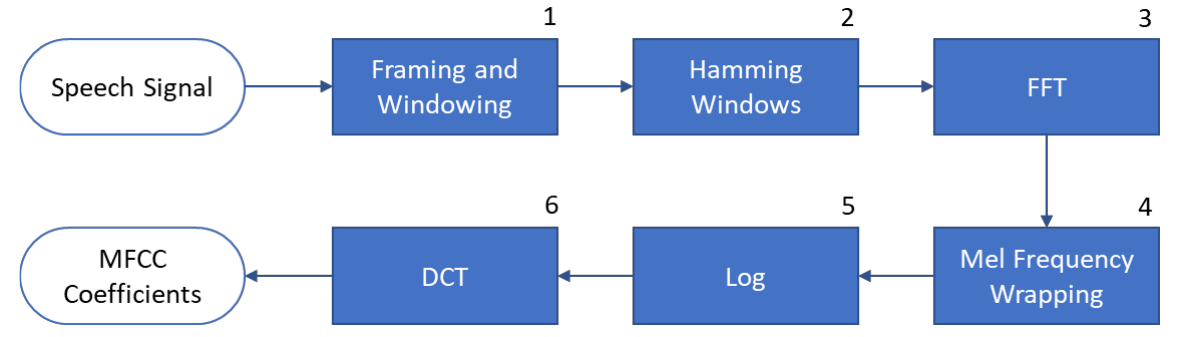}
\caption{Block diagram of MFCC feature extraction (adapted from \cite{b44}.}\label{fig4}
\end{figure*}

Feature extraction is the first step for an ASR system. It converts the waveform speech signal to a set of feature vectors with the aim of having high discrimination between phonemes \citep{b31}. The feature extraction performs all the required measurements on the selected segment that will be used to make a decision \citep{b32}. The measured features may be used to update long-term statistical measures to facilitate the adaptation of the process to varying environmental conditions (mainly the background) \citep{b33}. Feature extraction will determine the voice areas in the recording to be written out and extract sequences of acoustic parameters from them. There are many techniques for feature extraction, as reported in  \citep{b34}, including:
\begin{itemize}
    \item Linear Predictive coding (LPC) \citep{b35}: LPC is a technique for signal source modelling in speech signal processing. 
    \item RelAtive SpecTral (RASTA) filtering \citep{b36}: RASTA is designed to decrease the impact of noise as well as heighten speech. This technique is widely used for noisy speech.
    \item Linear Discriminant Analysis (LDA) and Probabilistic LDA \citep{b37}: This technique uses the state-dependent variables of Hidden Markov Model-based (HMM) on i-vector extraction. The i-vector is a low dimensional vector with a fixed length that contains relevant information.
    \item Mel-frequency cepstrum (MFCCs): It is the most commonly used technique, with a frameshift and length usually between 20 and 32 ms, using 1024 frequency bins, 26 mel channels and between 10 and 40 cepstral coefficients with cepstral mean normalisation \citep{b46,b58,b59,b60}. This technique has low complexity and a high ACC of recognition. Mel-Frequency Cepstrum Coefficient (MFCC) is the usual method for character extraction in most papers tackling the design of speech recognition systems for limited vocabulary \citep{b33,b42,b43}. The public sphinx base library provides an implementation of this method that can be used directly, as was done in \citep{b44}. Figure~\ref{fig4} provides a brief description of the MFCC method, encompassing six steps as described below. 
\end{itemize}

\begin{figure*}
\centering
\includegraphics[width=8cm]{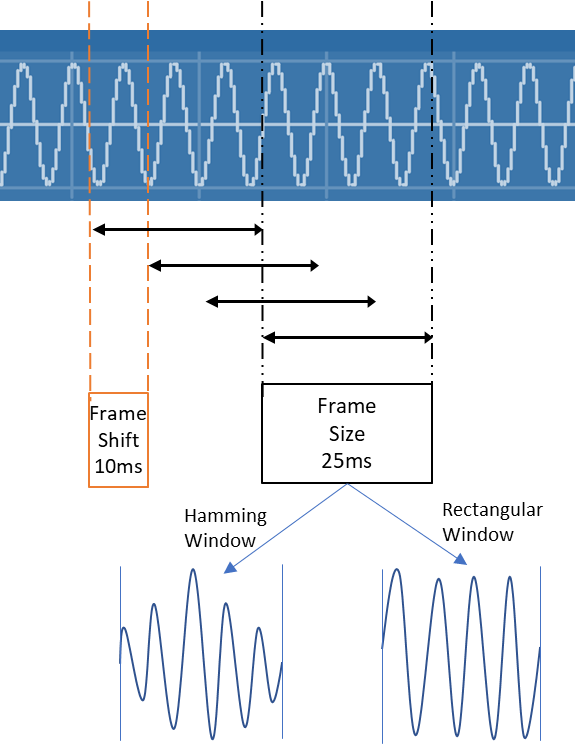}
\caption{Illustration of Framing and Windowing process with a Frame Shift of 10ms, a Frame size of 25ms and two representations: Hamming window and Rectangular window. }\label{fig9}
\end{figure*}

\begin{enumerate}
    \item Framing and windowing: For the acoustic parameters to be stable, the speech signal must be examined over a sufficiently short period. This step aims to cut windows of 20 to 30 ms. The process is illustrated in  Figure~\ref{fig9}.
    \item Hamming window: The Hamming window is used to reduce the spectral distortion of the signal. This is in contrast to the Rectangular window, which is simple but can cause problems since it slices the signal boundaries abruptly; the Hamming window reduces the signal values toward zero at the boundaries. This can help avoid discontinuities. This is expressed mathematically as follows: 
    \begin{equation}
        y(n)=x(n)\times w(n),~0\leq n\leq N-1,
    \end{equation}
    where $n$ is the number of windows, $N$ is the number of samples in each frame, $y(n)$ the output signal, $x(n)$ the input signal and $w(n)$ the Hamming windows defined as:
    \begin{equation}
        w(n) = 0.54 -0.46\cos\left(\frac{2 \pi n}{N} \right).
    \end{equation}
    \item	Fast Fourier Transformation (FFT) does the conversion of time domain windows into the frequency domain by discretising and interpolating the window onto a regular grid before the Fourier transform is applied. The FFT decreases the computation requirements compared to the discrete Fourier transform. The latter is defined as: 
\begin{equation}
    S(k)=\sum_{n=0}^{N}y(n)\text{e}^{-2\pi K\frac{n}{N}},~0\leq k\leq N,
\end{equation}
where $N$ is the number of windows, $k$ is the index of the coefficient and $K$ is the number of coefficients. 
\item	Mel Filter Wrapping banks allow reproduction of the selectivity of the human auditory system by providing a coefficient that gives the energy of the signal: 
\begin{equation}
    X(m)=\sum_{k=0}^{N-1}|S(k)|W(k,m),~1\leq m \leq M, ~M<N,
\end{equation}
where $m$ is the index of the filter, $M$ is the number of filters and $W$ is the weight function with inputs. The $k^{th}$ energy spectrum bin contributing to the $m^{th}$ output band.
\item Log allows one to obtain the logarithmic spectrum of Mel and to compress the sum $X(m)$ using:
\begin{equation}
    X^\prime(m)=\ln(X(m)).
\end{equation}
\item	Discrete Cosine Transform (DCT) reduces the influence of low-energy components. MFCC coefficients are obtained by the discrete cosine transform given by:
\begin{equation}
    c(k)=\sum_{m=0}^{N-1}a_m \cos\bigg[\frac{\pi}{M}(m+\frac{1}{2})k\bigg]X^\prime(m),~0\leq k\leq K,
\end{equation}
 where $a_0=\frac{1}{\sqrt{M}}, a_m=\sqrt{\frac{2}{M}}$.
\end{enumerate}
After this last step, the coefficients are returned as outputs.

\subsection{Acoustic modelling}
Acoustic modelling (AM) of speech typically describes how statistical representations of the feature vector sequences computed using the speech waveform are established. It aims to predict the most likely phonemes in the audio that has been given as input \citep{b2}. Challenging configurations mainly involving noise may occur. Two cases can be identified, namely the noise is isolated to a single band of frequencies, or the noise is not isolated, meaning it is on several bands. In the first case, AMs are still reliable enough to make the right decision.  In the second case, models have no good information and they are prone to mistakes \citep{b45}. The Bayesian process applied to ASR is as follows: Assuming that $x$ is a sequence of unknown acoustic vectors and $w_i~(i=1,\cdots, K)$ is one of the $K$ possible classes for an observation, the recognised class is given by: 
\begin{align}
     w^*&=\arg_i \max \frac{P(x|w_i)P(w_i)}{P(x)}\\
     &=\arg_i \max P(x|w_i)P(w_i).
\end{align}
The recognised word $w^*$ will therefore be the one that maximises this quantity, among all the candidate words $w_i$.  
The probability $P(x|w_i)$ of observing a signal, $x$, knowing the sequence needed for the AM estimation. The a prior  $P(w_i)$ of the sequences is independent of the signal and needs an LM for estimation. $P(X)$ is the probability to observe the sequence of acoustic vectors $X$. It is the same for each phoneme sequence (because $P(X)$ does not depend on $W$), so it can be ignored \citep{b46}.   

For ASR using limited vocabulary, the most commonly used AM to estimate the probability $P(X|W)$ is the HMM \citep{b42,b44,b47,b138}. The HMM is considered a generator of acoustic vectors. It is a finite-state automaton in which the transition from state $q_i$ to state $q_j$ has a probability of $a_{ij}$ at each time unit. This transition generates an acoustic vector $x_t$ with a probability density $b_j (x_t)$. The HMM is then given by $M= (\pi_i,A,B)$, where $\pi_i$ represents the initial probability distribution, $A$ is the transition probability matrix $a_{ij}$, and $B$ is the set of observation densities $B=\{ b_j (x_t)\}$. Several approaches for learning HMM models have been proposed, such as the maximum likelihood estimation \citep{b49} and forward-backward estimation \citep{b50}.

\subsection{Language model }
The LM in ASR is used to predict the most likely word sequence for a given text. In limited vocabulary, the training text is divided into word classes during the training phase. Based on word classes, the class-based n-gram LM is elaborated on. Thereafter, the standard bigram model, the class-based bigram model, and the interpolated model are obtained and used by the speech recognition system \citep{b51}. 

The language statistical model (for a sequence of words $W=w_1,w_2,\cdots,w_N$) consists in calculating the probability $P(W)$:
\begin{align}
    P(w)&=\prod_{i=1}^{N} P(w_i|w_1,\cdots,w_{i-1})\\
        &=\prod_{i=1}^{N}P(w_i|h_i),
\end{align}
    where $h_i=w_1,...,w_{i-1}$ is considered as the history of the word $w_i$ and $P(w_i |h_i)$ is the probability of the word $w_i$, knowing all the previous words. 

In practice, as the sequence of words $h_i$ becomes richer, an estimation of the values of the conditional probabilities $P(w_i |h_i)$ becomes more and more difficult because no corpus of learning text can observe all possible combinations of $h_i=w_1,...,w_{i-1}$.

To reduce the complexity of the LM, and consequently of its learning, the n-gram approach can be used. The principle is therefore the same and only the history is limited to the previous $n-1$ words. The probability $P(w)$ is thus approximated as:
\begin{align}
    P(w)&=\prod_{i=1}^{N} P(w_i|w_{i-n+1},\cdots,w_{i-1}).
\end{align}
It is possible to find the probability of the occurrence of a word $w_i$ in the learning corpus:
\begin{align}
    P(w_i)=\frac{C(w_i)}{C},
\end{align}
where $C(w_i)$ is the number of times the word $w_i$ has been observed in the learning corpus and C the total number of words in the corpus.
In practice, depending on the size of the learning corpus, different sizes of the history can be chosen. We then speak of a unigram model if $n = 1$ (without history), a bigram if $n = 2$ or a trigram if $n = 3$.\\

The N-gram language model Pocket sphinx-based is used to express syntactic constraints between words. For automatic speech recognition for limited vocabulary, the online tool LMTool of CMU is recommanded to train voice data on the network server \citep{b44,b148,b138}. LMTool use a corpus in the form of an ASCII text file to make appear language model and dictionnary. Always for ASR limited vocabulary, many authors use the HTK decoding parameters in language model scale factor \citep{b137,b81,b144,b143,b112,b142}.

\subsection{Pronunciation model}
The pronunciation dictionary models describe how a word is pronounced and represented. For small vocabulary, word-based models are used to define whole words as individual sound units. The pronunciation dictionary is a part of the pronunciation model. Translation of speech signal into text is achieved by classifying the speech signal into small sound units. The pronunciation model then determines how these small units can be combined to form valid words. For limited vocabularies, pronunciation models can be constructed by using handwritten word pronunciations, deriving them with phonological rules, or finding frequent pronunciations in a hand-transcribed corpus \citep{b52}. Another way to design the pronunciation model for limited vocabularies is to develop independent statistical models for each word in the dataset. The idea is to design a system in which each word has a number of parts, and then a model is trained to recognise each part of the word \citep{b53}.
Once the acoustic, the language, and the pronunciation models are developed, the decoder uses them to output the text corresponding to the received audio signal.

\subsection{Decoder}
A decoder is seen as a graph search algorithm which combines acoustic and linguistic knowledge to automatically transcribe the input record \citep{b2}. The goal of decoding is to deduce the sequence of states that generated the given observations. From this sequence of states, it is easy to find the most likely sequence of phonemes that matches the observed parameters. For a limited vocabulary, the Viterbi search algorithm \citep{b54} uses the probabilities of the AM and those of the LM to accomplish the decoding task. The Viterbi decoder is good for short commands, meaning for small or limited vocabulary \citep{b55,b56}. It seeks the most probable candidate among states in an HMM. This search is performed given the probability of observations obtained from the AM, for each time step for each of the states (cepstral coefficient vector corresponding to the time step). 

\section{Automatic speech recognition approaches}

Three techniques in artificial intelligence are used for ASR in general, namely machine learning (ML), DL, and deep reinforcement learning (DRL).   
  
  \subsection{Machine Learning}
Machine learning (ML) is an artificial intelligence technique that refers to systems that can learn by themselves \citep{b57}. ML implies teaching a computer to recognise patterns in contrast to the traditional approach, which consists of programming a computer with specific rules. The teaching is done through a training process that involves feeding large amounts of audio data to the algorithm and allowing it to learn from data and detect patterns that can later be used to achieve some tasks \citep{b58,b59}. ML techniques can be grouped into four categories \citep{b60}:
\begin{itemize}
    \item Supervised Learning consists of inferring a classification or regression from labelled training data, to develop predictive models. Classification techniques predict discrete responses while regression techniques predict continuous responses.
    \item Unsupervised Learning consists of drawing inferences from datasets composed of unlabelled input data. The most common technique in this category is clustering, which is used for exploratory data analysis in order to find hidden patterns or groupings in data.
    \item Semi-supervised Learning. The training of the system in this learning technique makes use of both labelled and unlabelled data. This type of training is adequate when it is very expensive to obtain labelled data.
    \item Active Learning. This approach is used when there is a lot of unlabelled data, but labelling is expensive. The algorithm interactively queries the user to label data.
\end{itemize}

The different ML steps for speech recognition are provided in Figure~\ref{fig5} and are detailed as follows:

\begin{figure*}
\centering
\includegraphics[width=16cm]{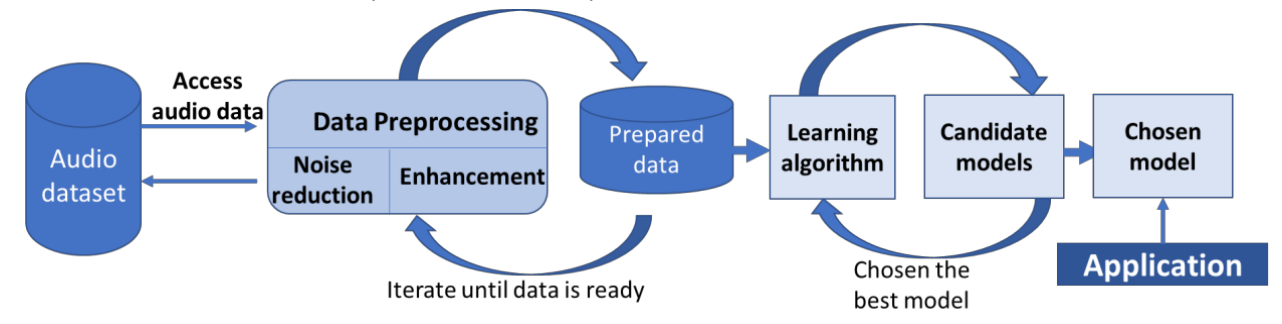}
\caption{ML technique}\label{fig5}
\end{figure*}

\begin{enumerate}
    \item The first step is to select and prepare a training dataset composed of audios (from words or sentences) that have been acquired through microphones. This data will be used to feed the ML model during the learning process so that it can determine texts corresponding to audio inputs. Data must be meticulously prepared, organised and cleaned with the aim to mitigate bias during the training process. 
    \item The second step is to perform a pre-processing on the input data. This pre-processing includes the reduction of the noise in the audio and the enhancement of data.
    \item The third step consists of choosing a parametric class where the model will be searched.  This is a fine-tuned process that is run on the training dataset. For speech recognition using limited vocabulary, some algorithms such as the Maximum Likelihood Linear Regression algorithm \citep{b44} can be used to train the AM, and the Viterbi algorithm \citep{b43, b44} for decoding. The type of algorithm to use depends on the type of problem to be solved.
    \item The fourth step is the training of the algorithm. This is an iterative process. After running the algorithm, the results are compared with the expected ones. The weights and biases are eventually tuned  via the back propagation optimisation, to increase the accuracy of the algorithm. This process is repeated until a certain criterion is met and the resulting trained model is saved for further analysis with the test data.
    \item The fifth and final step is to use and improve the model. The model is then used on new data. 
\end{enumerate}
Different ML methods  have been used for acoustic modelling in speech recognition systems \citep{b60}. The evaluation, decoding and training of HMMs are done by ML forward-backward \citep{b50}, Viterbi \citep{b54} and Baum-Welch algorithms \citep{b49}, respectively. In their work, Padmanabhan et al. \citep{b60} review these methods.

\subsection{ Deep Learning}
Deep Learning (DL) is a set of algorithms in ML. It uses model architectures made up of multiple non-linear transformations (neural networks) to model high-level abstractions in data \citep{b61}. Deep Neural Networks (DNNs) work well for ASR when compared with Gaussian Mixture Model-based HMMs (GMM-HMM) systems, and they even outperform the latter in certain tasks \citep{b62}. DL employs the Convolutional Neural Network (CNN) approach which owns the ability to automatically learn the invariant features to distinguish and classify the audio \citep{b63}.

By learning multiple levels of data representations, DL can derive higher-level features from lower-level ones to form a hierarchy. For instance, in a speech classification task, the DL model can take phoneme values in the input layer and assign labels to the word in the sentence in the output layer. Between these two layers, there are a set of hidden layers that build successive higher-order features that are less sensitive to conditions, such as noise in the user’s environment \citep{b64}.

\begin{figure*}
\centering
\includegraphics[width=16cm]{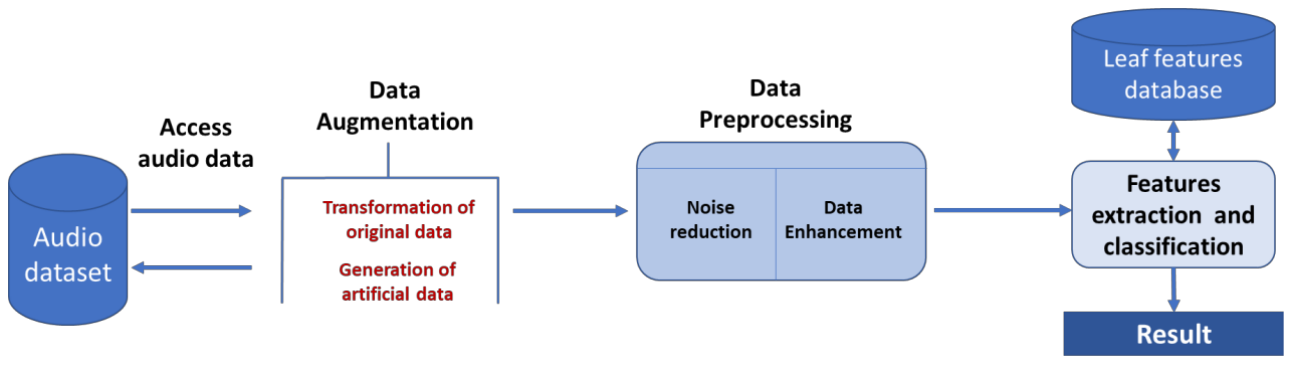}
\caption{Steps of DL}\label{fig6}
\end{figure*}

DL can be implemented using various tools. However, Tensor Flow seems to be one of the best application methods currently available \citep{b65}.
Figure~\ref{fig6} gives the steps of DL in ASR. Data augmentation helps to improve the performance of the model by generalising better and thereby reducing overfitting \citep{b66, b67}. Data augmentation creates a rich, diverse set of data from a small amount of data. Data augmentation can be applied as a pre-processing step before training the model or later, directly in real-time. Different augmentation policies can be applied to audio data such as Time warping, Frequency masking, and Time masking. Recently, a new augmentation method called SpecAugment has been proposed by Park et al. in \citep{b68} for the ASR system. They combined the warping of the features and the masking of blocks of frequency channels, as well as the blocks of time steps. To ease the augmentation process, a recent free MATLAB  Toolbox called Audiogmenter\footnote{https://github.com/LorisNanni/Audiogmenter.}  has been proposed \citep{b69}. 

The feature extraction process aims to remove the non-dominant features and therefore reducing the training time while mitigating the complexity of the developed models.

 \subsection{ Deep Reinforcement Learning}
A speech recognition system can be vulnerable to a noisy environment. To address this issue, deep reinforcement learning (DRL) can achieve complex goals in an iterative manner, which makes it suitable for such applications. Reinforcement learning is a popular paradigm of ML, which involves agents learning their behaviour by trial and error. DRL is a combination of standard reinforcement learning with DL to overcome the limitations of reinforcement learning in complex environments with large state spaces or high computation requirements. DRL enables software-defined agents to learn the best actions possible in virtual environments to attain their goals \citep{b70}. This technique has recently been applied to limited vocabulary such as the “Speech Command” dataset in \citep{b71} or larger vocabulary such as \citep{b72}.
Regardless of the artificial intelligence technique that is used, an important prerequisite remains, namely the dataset.

\section{Datasets management}
\subsection{Construction}
Speech recognition research has traditionally required the resources of large organisations, such as universities or corporations \citep{b14}. Microphones can save data coming from multiple sources; this permits the collection of enough data for recognition. Although this approach remains the most frequent use case, it is subject to some challenges including speaker localisation, speech enhancement, and ASR in distant-microphone scenarios \citep{b73, b74, b75, b76, b77, b78}.

\begin{figure*}
\centering
\includegraphics[width=16cm]{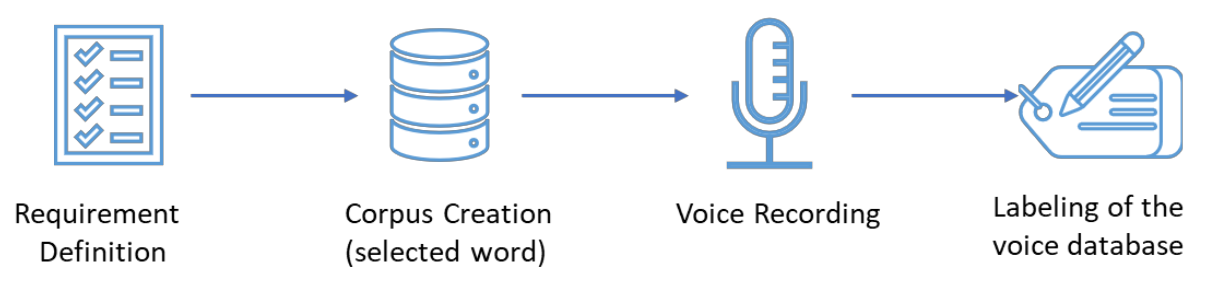}
\caption{Construction of dataset for limited vocabulary}\label{fig7}
\end{figure*}

Datasets with limited vocabulary may be  a subset of larger speech datasets \citep{b79}. Datasets are used to train and test ASR engines. It is important to have a multi-speakers database \citep{b40,b42}. To avoid inconsistencies between datasets, the collection is done during a short period (the same day if possible) \citep{b80}. For the best results, the corpus of acoustic data used for learning must be performed in a good quality recording studio, but it does not have to occur in a professional studio. For instance, authors in\citep{b47} used a Handy Recorder with four channels (H4n) that is  affordable\footnote{https://www.amazon.com/Zoom-Four-Channel-Handy-Audio-Recorder/dp/B07X8CJBW9}. The creation of an audio dataset goes through four steps as shown in Figure~\ref{fig7}. These steps are: requirement definition, corpus creation, voice recording, and labelling of the voice database.

\begin{enumerate}[label=\alph*.)]
  \item Requirement definition:
The first step for the construction of a dataset for limited vocabulary consists of selecting a suitable environment, which can be a room with closed doors to reduce noise. The use of studio recorded audio is unrealistic, as these audios are free of background noise and recorded with high-quality microphones and are in a formal setting. Good ASR models should work equally well in moderated noisy environments with natural voices. For this, the audio recordings should be made in a simple environment with closed doors.
	For good recognition, audio recordings should be made with several people (female and male) having different tones. The data should have a fixed and short duration to facilitate the training of the learning model and the evaluation process \citep{b14}.
\item Corpus creation: 
After requirement definition, words or sentences of the limited vocabulary are chosen. The choice is made according to the needs, and it is limited to the context in which speech recognition will be performed. Only necessary data should be selected. The maximum size of the corpus for limited vocabulary is generally around a thousand of words.
\item Voice recording:
Sounds are recorded using a microphone that matches the desired conditions. It should minimise differences between training conditions and test conditions. Speech recordings are generally performed in an anechoic room and are usually digitised at 20 kHz using 16 bits \citep{b81, b79} or at 8 kHz \citep{b47} or 16 kHz \citep{b14, b80}. The waveform audio file format container with file extension .wav is generally used \citep{b14, b79}. The WAV formats encoded to Pulse-Code Modulation (PCM) allow one to obtain an uncompressed and high-fidelity digital sound. Since these formats are easy to process in the pre-processing phase of speech recognition and for further processing, it is necessary to convert the audio files obtained after the recording (for instance OGG, WMA, MID, etc.) into WAV format.
\item 	Labelling of the voice database:
Most ML models are done in a supervised approach. For supervised learning to work, a set of labelled data from which the model can learn to make correct decisions is required. The labelling step aims to identify raw data (text and audio files) and to add informative labels in order to specify the context so that an ML  model can learn from it. Each recorded file is marked by sub-words or phonemes. Labelling indicates which phonemes or words were spoken in the audio recording. To label the data, humans make judgments about some aspects of the unlabelled audio file. For example, one might ask to label all audios containing a given word or phoneme.
	The ML model uses human-supplied labels to learn. The resulting trained model is used to make predictions on new data. Some software such as Audio Labeller\footnote{https://www.mathworks.com/help/audio/ref/audiolabeler-app.html}  allows one to define reference labels for audio datasets. Audio Labeler also allows one to visualise these labels interactively.
\end{enumerate}

The minimal structure of a dataset for limited vocabulary takes into account elements such as the path to the audio file, the text corresponding to the audio file, the gender of the speaker, the age and the language used if there are many languages in the dataset\footnote{https://huggingface.co/datasets/timit\_asr }. An excerpt of such a dataset for ASR using limited vocabulary is given in Table \ref{tab3}.

\begin{table*}
\caption{Structure of a multilingual dataset for limited vocabulary}\label{tab3}
\centering
\begin{tabular}{ p{3cm} p{3cm} p{3cm} p{3cm} p{3cm} }
 \hline
 Audio Wav file&	Text&	Age&	Sex	&Accent (Language)\\
 \hline
C:/dataset/audio$_1.wav$&	One&	23	&M&	English\\

C:/dataset/audio$_2.wav$&	Two&	20&	F&	English\\

$\cdots$&$\cdots$&$\cdots$&$\cdots$&$\cdots$\\

C:/dataset/audio$_n.wav$&	Dix	&35	&M&	French\\
\hline
\end{tabular}

\end{table*}

If there are several speakers, then the speaker's index will be one key element in Table \ref{tab3}. Some researchers also take into account the emotion of the speaker, mentioning if he is neutral, happy, angry, surprised or sad \citep{b82}. 

\subsection{ Languages and Datasets}
In the literature, several works have developed datasets such as VoiceHome2 \citep{b83} that developed a French corpus for distant microphone speech processing in domestic environments. Also regarding French, the work in \citep{b84} proposes a multi-source data selection for the training of LMs dedicated to the transcription of broadcast news and TV shows.
An important work focusing on the English language is the reduced voice command database \citep{b85}. It has been created from a worldwide cloud speech database and in combination with training, testing and real-time recognition algorithms based on artificial intelligence and DL neural networks. Another important English dataset for digits from 0 to 9, and 10 short phrases is the AV Digits Database \citep{b86}. In a survey, 53 participants consisting of 41 males and 12 females were asked to read digits in English in random order five times. In another study, 33 agents were asked to record sentences, each sentence was repeated five times in 3 different modes: neutral, whisper and silent speech. A  larger dataset is the Isolet dataset proposed in \citep{b87}. This dataset contains  150 voices divided into five groups of 30 people. In this dataset, each speaker pronounces  each letter of the English alphabet twice, which provides a set of 52 training examples for each speaker.

Apart from French and English, other languages, such as Sorani Kurdish have been tackled. BD-4SK-ASR (Basic Dataset for Sorani Kurdish Automatic Speech Recognition) is an experimental dataset which is used in the first attempt in developing an ASR system for Sorani Kurdish \citep{b88}.

A very large project run by Mozilla is the Common Voice Project initiated with the aim of producing an open-source database for ASR \citep{b89}. The Mozilla Common voice dataset created in 2017 is intended for developers of language processing tools. In November 2020, more than 60 languages were represented on the platform. These languages include French, English, Chinese, Danish and Norwegian.
It is very important that speech recognition systems be tested for efficiency, regardless of the language.

\section{Performance metrics}
The quality of the output transcript of an ASR system is traditionally measured by the word error rate (WER) metric:
\begin{equation}
    WER=\frac{S+I+D}{N},	
\end{equation}
where $S$ is the number of incorrect words substituted, $I$ is the number of extra words inserted, $D$ is the number of words deleted and $N$ is the number of words in the correct transcript. The WER metric is estimated in terms of percentage. However, it is important to note that it is possible to have a WER value exceeding 100\% and the WER threshold for acceptable performance depends on the applications. However, Johnson et al. \citep{b90} have shown that it is still possible to get a good retrieval performance with a WER value up to 66\%. However, the precision begins to fall off quickly when WER gets above 30\% \citep{b90}. Having an estimation of WER value and knowing the threshold for usable transcripts, the allocation of processing resources can be oriented only to those files predicted to yield usable results. If a spoken word is not included in the vocabulary, the corresponding transcript will be considered as an error and the WER value will increase.

For detection of keywords,  authors in \citep{b91} proposed a new method to calculate the mean of accuracies of each keyword ($acc_i$). This is computed depending on the number of keywords correctly predicted ($N_{cp}$), the number of keywords not yet predicted ($N_{ny}$) and the number of keywords incorrectly predicted ($N_{ip}$):
\begin{equation}
    acc_i=\frac{N_{cp}-N_{ip}}{N_{cp}+N_{ny}}.
\end{equation}
The ACC is computed from the above as:
\begin{equation}
   ACC=\frac{1}{N} \sum_{i=0}^{N-1}acc_i. 
\end{equation}

The WER metric works well when the morphology of the language is simple \citep{b8}. Otherwise, more adequate metrics should be applied, such as the Letter or Character Error Rate (LER or CER) \citep{b92, b93}, the Syllable Error Rate (SylER) \citep{b94} or Speaker Attributed Word Error Rate (SA-WER) \citep{b95}.\\
The use of OPD (Output Probability Distributions) and secondary classification has been proposed as a solution to improve the accuracy of ASR for isolated word in limited vocabulary \citep{b164}. To do this, it models the relationship between words. OPD represents the distribution of logarithmic probability of HMM’s set. For each word of vocabulary, a HMM is trained, an utterance is transfered to HMM and logarithmic probabilities are concatenated to give OPD.

The LMs can be evaluated separately from the AM and the most commonly used measure is perplexity \citep{b96}. This measure is calculated on a text not seen during training (test-set perplexity). Another measure is based on Shannon's game \citep{b97}. 
Existing frameworks are used to develop ASR systems. Comparing these and previous works can enable newcomers in the field to develop ASR systems. 

\section{Speech recognition frameworks or Toolkits}
Several toolkits for ASR have been developed to train datasets. Some toolkits are open-source code and others are not and their presentation is the aim of this section.

\subsection{Closed-source code systems}
Several closed-source systems are available for ASR, namely the Dragon Mobile software development kit (SDK), Google Speech Recognition API, SiriKit, Yandex SpeechKit and Microsoft Speech API \citep{b98}.

The Dragon Mobile SDK\footnote{https://www.nuance.com/dragon/for-developers/dragon-software-developer-kit.html },  developed by Nuance since 2011, provides speech services to enhance applications with speech recognition and text-to-speech functionality. It consists of a set of sample projects, documentation, and a framework to ease the integration of speech services into any application. It is a trialware that requires a paid subscription after 90 days. Popular mobile platforms are supported (Android, iOS, and Windows phone). It has been used in some works like \citep{b99} where a custom phonetic alphabet has been optimised to facilitate text entry on small displays.

The Yandex SpeechKit\footnote{https://cloud.yandex.com/en/services/speechkit }  can be used to integrate speech recognition, text-to-speech, music identification, and Yandex voice activation into Android mobile applications. The Yandex SpeechKit supports the following languages for speech recognition and text-to-speech: Russian, English, Ukrainian and Turkish. Yandex SpeechKit has been recently used in \citep{b100}. 

SiriKit\footnote{https://developer.apple.com/documentation/sirikit}  is a toolkit that allows one to integrate Siri into a third-party iOS or macOS application, to benefit from the ASR capabilities. Siri is a speech recognition personal assistant unveiled by Apple in 2011 and only works on iOS and macOS \citep{b101}. Even though potential dangers and ethical issues have been reported \citep{b102, b103}, SiriKit is used in numerous works including \citep{b104, b105}.

The Speech Application Programming Interface\footnote{https://docs.microsoft.com/en-us/previous-versions/windows/desktop/ms720151(v=vs.85)} (SAPI) developed by Microsoft enables developers to integrate speech recognition and speech synthesis within Windows applications \citep{b106}. Several versions of the API have been released either as part of a Speech SDK or as part of the Windows Operating System. SAPI is used in Microsoft Office, Microsoft Agent and Microsoft Speech Server. Several research works have also made use of this API \citep{b107, b108}.

Google Speech Recognition API\footnote{https://cloud.google.com/speech-to-text?hl=en }  is a C\# toolkit based on a model trained with English and 120 other languages. Google has improved its speech recognition by using a DNN in its applications, reaching an 8\% error rate in 2015 compared to 23\% in 2013 \citep{b109}. 

Table \ref{tab4} provides a summary of selected closed-source code speech recognition toolkits.

\begin{table*}
\caption{A non-exhaustive list of closed-source code speech recognition frameworks/toolkits}\label{tab4}
\centering
\begin{tabular}{ p{3cm} p{2cm} p{3cm} p{4cm} p{4cm}}
 \hline
Toolkit/$1^{st}$ Release &Programming Language& Targeting language& Applications& Applied technology\\
 \hline
Microsoft Speech API, 1995& C\#& English, Chinese,Japanese& Speech recognition, speech synthesis within windows applications&
Context-dependent DNN, HMM\\

 SiriKit, 2011&Objective-C&English, German, French& Speech recognition & DNN and ML\\
 
 Dragon Mobile SDK, 2011& C++&English, Spanish, French, German, Italian, Japanese& Speech recognition and text to speech&Markov processes\\
 
 Yandex Speechkit, 2014& C, Python&Russian, English, Ukrainian, Turkish&Speech recognition, text to speech, music identification and voice activation&Neural network\\
 
 Google Speech Recognition API, 2016& C\#&English, more than 120 languages&Speech to text, speech recognition & DL neural network\\
 \hline
\end{tabular}

\end{table*}

As conventional closed-source software, these toolkits are not flexible and are limited to the languages they have been designed for. Such limitations fostered the research community to develop open-source frameworks and toolkits.

\subsection{Open-source code systems}
A plethora of open-source frameworks, engines, or toolkits for ASR systems are proposed in the literature. The following is a non-exhaustive list of main works or projects. Most of them can easily handle small and large vocabulary.  
The HTK\footnote{https://htk.eng.cam.ac.uk/} is one of the first speech recognition software developed since the late 1980s. It has been available for the research community since 2000 and is maintained at the Speech Vision and Robotics Group\footnote{http://mi.eng.cam.ac.uk} of the Cambridge University Engineering Department (CUED) \citep{b110}. It provides recipes to build baseline systems with HMM. HTK is considered a very simple and effective tool for research \citep{b111, b112}. It can build a noise-robust ASR system in a moderated noisy level environment, especially for small vocabulary systems. It is a practical solution to develop fast and accurate Small Vocabulary Automatic Speech Recognition (SVASR) \citep{b113}. Several works use this toolkit \citep{b114, b115, b116}.
One of the most popular toolkits is the CMU Sphinx\footnote{https://cmusphinx.github.io/ }, designed for both mobile and server applications. CMU Sphinx is in fact a set of libraries and tools that can be used to develop speech-enabled applications. It is developed at Carnegie Mellon University in the late 1980s \citep{b117}. Several versions have been released including Sphinx 1 to 4, and PocketSphinx for hand-held devices \citep{b118}. CMU Sphinx is currently attracting the attention of the research community. It offers the possibility to build new LMs using its language Modelling Tool\footnote{http://www.speech.cs.cmu.edu/tools/lmtool-new.html }.

A former toolkit that may no longer be available is the Rapid Language Adaptation Toolkit (RLAT) introduced in \citep{b119} and used in \citep{b120, b121}. The website for the project is, unfortunately, no longer available. 
Kaldi is an extendable and modular toolkit for ASR \citep{b122}. The large community behind the project provides numerous third-party modules that can be used for several tasks. It supports DNN and offers excellent documentation on its website\footnote{https://kaldi-asr.org/}. Several works have been based on this toolkit including \citep{b123,b124,b125}. 

Microsoft also proposes an open-source Cognitive Toolkit\footnote{https://docs.microsoft.com/en-us/cognitive-toolkit/} (CNTK). It is used to create DNNs that power many Microsoft services and products. It enables researchers and data scientists to easily code such neural networks at the right abstraction level, and to efficiently train and test them on production-scale data \citep{b126,b127}. Although it is still used, it is a deprecated framework.

Julius\footnote{https://julius.osdn.jp/en\_index.php} is software normally designed for LVCSR. It is based on word $N-gram$ and context-dependent HMM. It is used in several works such as \citep{b128,b129}. 

Simon toolkit\footnote{https://simon.kde.org/ } is a general public license speech recognition framework developed in C++. It is designed to be as flexible as possible and it works with any language or dialect. Simon makes use of KDE libraries, CMU SPHINX or Julius together with HTK and it runs on Windows and Linux.

Praat\footnote{https://www.fon.hum.uva.nl/praat/} is a framework that enables speech analysis, synthesis, and manipulation. In addition, it allows speech labelling and segmentation. It has been used in \citep{b130, b131}.

Mozilla Common Voice\footnote{https://commonvoice.mozilla.org/ 
} is a free speech recognition software for developers that can be integrated into projects. It works with DNN technology and targets several languages \citep{b89}. Besides Common Voice, Mozilla has also developed DeepSpeech\footnote{https://github.com/mozilla/DeepSpeech}, an open-source Speech-To-Text engine. It makes use of a model trained by the ML techniques proposed in \citep{b132}. 

Another larger project is the OpenSMILE (open-source Speech and Music Interpretation by Large-space Extraction) project\footnote{https://www.audeering.com/opensmile/} that is completely free to use for research purposes. It received a lot of attention from the research community and claims more than 150,000 downloads. 
A recent model called Jasper (Just Another Speech Recognizer) has been introduced in 2019 \citep{b133}. It can be used with the OpenSeq2Seq\footnote{ttps://nvidia.github.io/OpenSeq2Seq/html/index.html} TensorFlow-based toolkit. OpenSeq2Seq enables, among others, speech recognition, speech commands and speech synthesis. It is used in recent works such as \citep{b134, b135}.

Other recent engines for ASR have been released such as Fairseq\footnote{https://github.com/pytorch/fairseq} and Wav2Letter++\footnote{https://github.com/facebookresearch/wav2letter} (both developed by Facebook), Athena\footnote{https://github.com/athena-team/athena}, ESPnet\footnote{https://espnet.github.io/espnet/}, and Vosk\footnote{https://alphacephei.com/vosk/} which is an offline ASR toolkit. Table \ref{tab5} provides a summary of some open-source speech recognition toolkits.

\begin{table*}
\caption{A non-exhaustive list of open-source speech recognition frameworks/toolkits}\label{tab5}
\centering
\begin{tabular}{ p{3cm} p{2cm} p{3cm} p{4cm} p{4cm}}
 \hline
Toolkit / $1^{st}$ Release&Programming language&License&Trained models&
Applied technology\\
 \hline
Praat, 1991& C/C++, Objective C& GNU General Public License&-&Neural networks\\

HTK, 2000&C/C++&HTK Specific License&English&HMM, DNN\\

CMU Sphinx, 2001&Java&BSD License&English plus 10 other languages&HMM\\

Simon, 2008& C++&GNU General Public License&-&HMM\\

Kaldi, 2009&C++&Apache&Subset of English& DNN \\

OpenSmile, 2010&C++&Free for non-commercial use&-&Machine/DL\\

Wav2Letter, 2016&Torch (Lua)&BSD License&English& Neural Networks\\

Microsoft CNTK, 2016&C++&BSD License&-& DNN\\

Mozilla Common Voice, 2017&C, Python, Java&Public domain CC-O&English, French, German, Chinese, $\cdots$ & DNN\\

OpenSeq2Seq, 2018&Python&Apache&-& DNN\\

Fairseq, 2019&Python&BSD Licence&English, Chinese, German, and French&Neural Networks\\
 \hline
\end{tabular}

\end{table*}

Several works have been performed for limited vocabularies in ASR using the above toolkits. We make a summary of these works in the following section.

\section{Summary of related work on ASR using limited vocabulary}
\subsection{Selected works on ASR using limited vocabulary}
ASR using limited vocabulary has attracted the attention of researchers. The literature proposed systems for European languages such as English \citep{b136, b137, b138}, Estonian \citep{b81}, Norwegian \citep{b139}, Romanian \citep{b140, b51}, Macedonian \citep{b42}, as well as Asian languages including Chinese \citep{b44, b43}, Hindi \citep{b141}, Indonesian \citep{b113}, Arabic \citep{b142,b143}, Bangla  \citep{b144}, Malayalam \citep{b145,b146}, Tamil \citep{b147}, Marathi \citep{b111}, Punjabi \citep{b41} and Urdu \citep{b148,b149}. Table \ref{tab6} provides a list of selected works done on ASR using limited vocabulary base. 
\end{multicols}
\begin{landscape}
\begin{table*}
\caption{Summary of works on ASR using limited vocabulary.}\label{tab6}
\centering
\begin{tabular}{ p{5cm} p{3cm} p{4cm} p{2cm} p{2cm} p{2cm} p{4cm}}
 \hline
Papers, Year&Language&Toolkit/Model&Noisy&Acc (\%)&\# Speakers&Dataset \\
 \hline
\citep{b137}&English&HTK&Yes& -&10&780 utterances\\

 \citep{b140}&Romanian&LGB and LVQ3&Yes&95.33&30&10 Digits\\

\citep{b81}&Estonian&HTK&No& 97&60&400 utterances of numbers\\

 \citep{b147}&Tamil& Feedforward neural networks&Yes&81&10 children&20 phonemes\\

 \citep{b145}&Malayalam&From scratch using ANN&Yes&
89&-&5 words\\

 \citep{b143}&Arabic&HTK&Yes&81.79&59&944 sentences\\

\citep{b144}&Bangla&HTK&Yes&90-95&100&10 digits\\

\citep{b112}&Yoruba/ Hebrew&HTK&No&100&3&50 words/ short phrases\\

\citep{b148}&Urdu&CMU Sphinx&Yes&60&10&52 words\\

\citep{b142}&Arabic&HTK&No&97.99&13&33 words\\

 \citep{b146}&Malayalam&From scratch: DWT and ANN&Yes&80&1&250 words\\

\citep{b138}&English&CMU Sphinx&No&71.73&-&11 words\\

\citep{b43}&Chinese&From scratch and HMM-based&Yes&89.6&8&640 speech samples\\
 
\citep{b47}&Wolof&HTK&Yes&81.9&25&Digits\\

\citep{b141}&Hindi&Microsoft Speech Server&Yes&90&24&79 words\\

\citep{b42}&Macedonian&HTK&Yes&94.52&30&188 words\\

\citep{b44}&Chinese&PocketSphinx (CMU)&Yes&90&1&10 voice commands\\

\citep{b136}&English&Kaldi&Yes&46.4&20&256 words\\

\citep{b111}&Marathi&HTK&No&80-90&5&620 words\\

\citep{b51}&Romanian&SRILM toolkit  and ProtoLOGOS &No&59.57&30&762 phrases\\

\citep{b149}&English&Kaldi&Yes&98&-&62 keywords\\

\citep{b139}&Norwegian&CMU Sphinx&Yes&58.31&24&10 sentences\\

\citep{b113}&Indonesian&HTK&Yes&100&1&Digits\\

\citep{b152}&Amazigh&CMU Sphinx&No&92.22&30&13 commands\\

\citep{b41}&Punjabi&CMU Sphinx&No&69.6&50&91 words\\

\citep{b157}&Dutch&PPVT&Yes&75.97&132&384 words\\

\citep{b158}&Chinese&Keras/Tensorflow&Yes&97.5&-&877 words\\

\citep{b159}&English&Google's Speech-to-Text API&Yes&95.2&432&300 shorts phrases\\

\citep{b160}&Spanish&Kaldi&Yes&55.78&33&164 words\\

\citep{b161}&English&Labview and NI my RIO 1900&Yes&96.375&3&10 words\\

 \hline
\end{tabular}

\end{table*}
\end{landscape}

\begin{multicols}{2}
\subsection{Remarks}
Even though European and Asian languages are well represented in the selected works, we noticed a particular focus on well-resourced languages, namely English, Chinese, Romanian. Only a few works deal with African languages namely Wolof \citep{b47} and Yoruba \citep{b112}. The weak representation of African languages can be justified by the fact that most of them are under-resourced, in addition to the lack of local skills and awareness about the potential of ASR systems to prevent the extinction of under-resourced languages. 
Although most of the works did consider noisy environments, the severity was moderated and usually limited to natural environmental noises. Due to the limited size of the vocabulary, such noise levels have been easily mitigated. This justified the competitive ACC in most works.

\begin{figure*}
\centering
\includegraphics[width=7cm]{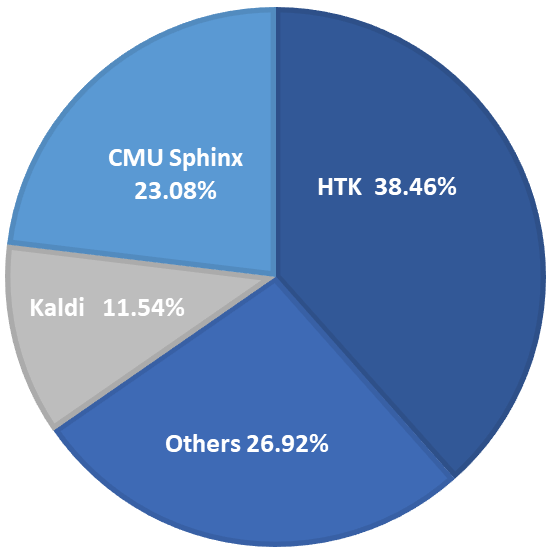}
\caption{The percentage of toolkits used in selected works on ASR using limited vocabulary.}\label{fig8}
\end{figure*}

Figure~\ref{fig8} gives the percentage use of the different toolkits. We observe that the HTK Toolkit is the most used for speech recognition with regards to limited vocabulary. Its success is explained by the fact that it eases the manipulation of HMM parameters for the training and testing stage of system development.

In general, the Bayesian equation allows one to find the probability that a word will be recognised in the speech recognition process for limited vocabulary, and in turn the sequence of words corresponding to a speech sequence. HMM and DNN are decent AMs that achieve good speech recognition results. HMM is used to account for variability in speech and DNN, with many hidden layers, have been shown to outperform GMMs on a variety of speech recognition benchmarks. The construction of datasets for ASR with limited vocabulary is done with speech recorded and digitised at varying frequencies between 8Khz and 20Khz in most cases. Finally, the ACC of the system is calculated by using the number of words correctly predicted, the number of words not yet predicted, and the number of words incorrectly predicted. ACC of ASR for limited vocabulary for previous works is between 46 and 100\%.

\section{Future directions}
The World has more than 7000 languages according to the Ethnologue website\footnote{https://www.ethnologue.com}, with the majority being under-resourced and even endangered. ASR systems offer an unprecedented opportunity to sustain such under-resourced languages and to fight the extinction of endangered ones. We should differentiate two types of under-resourced languages: those with an acknowledged written form and those without. In the first case, new datasets should be created and require new approaches for data recording and labelling, especially when there are not enough native speakers during the creation process. This is a challenge, especially with tonal languages. In fact, most languages in the developing world and especially in Sub-Saharan Africa are tonal \citep{b153}. In a  recent survey on ASR for tonal languages \citep{b154}, only two African languages were reported. In addition, some languages (especially dialects) share commonalities. Therefore, ASR systems with limited vocabulary targeting multiple similar languages can be designed. 

In the second case, when the language that does not have an acknowledged written form, a new approach should be designed. Normally, an ASR system aims to transcribe a speech into a text.  In ASR using a limited vocabulary, the transcript text is usually a command or a short answer that can be used by an application or system to perform an action. In this scenario, the system can be a combination of a Speech-to-Speech Translation (S2ST) and a Speech Recognition. First, the speech in a non-written language is directly translated into a speech in a well-resourced language such as English, then the transcript in a well-resourced language is retrieved (generated) and sent to the application. The Direct S2ST model has been developed in \citep{b155}, translating Spanish into English without passing through text. Their dataset is a subset of the Fisher dataset and is composed of parallel utterance pairs. The construction of datasets for ASR systems using limited vocabulary of nonwritten languages can also be based on the same principle.

Regardless of whether the language is written or not, more noise resistant models should be developed, because the available data for under-resourced languages could be of low quality.

The limited computing resources and poor internet connectivity in some regions can prevent the use of ASR systems. There is currently a shift of ASR models from the cloud to the edge. It is performed by reducing the size of models and making models fast enough so that they can be executed on typical mobile devices. The latest optimisation techniques to achieve this, such as pruning, efficient Recurrent Neural Network variants and quantisation, are presented in \citep{b156}. Although they provide remarkable results, such as reducing the size by 8.5x and increasing the speed by 4.5x, there is still a need to develop light and offline models that can be deployed on low-resource devices, such as off-the-shelf smartphones or raspberry Pi/Arduino modules.

\section{Conclusion}
This paper presented a review of ASR systems with a focus on limited vocabulary. After introducing the ASR principle in general and usual techniques used to perform recognition, this paper also discussed the management of datasets, the performance metrics to evaluate ASR systems, and toolkits to develop such systems. From the analysis of selected papers on ASR using limited vocabulary, HMM and DNN-based AMs achieve good speech recognition results. DNN even outperform GMMs on a variety of speech recognition benchmarks. Datasets with limited vocabulary are constructed with speech at frequencies between 8Khz and 20Khz. The evaluation of systems is mainly based on the ACC rather than the WER metric. Despite the satisfactory results, there is still much to do. In fact, developed systems deal mostly with well-resourced languages and most models are still running on servers. We hope that the ideas for future directions discussed regarding under-resourced/unwritten languages (investigating direct speech to speech translation) and designing “on the edge ASR systems for limited vocabulary” will draw the attention of the research community to develop systems especially for developing world, where the limitation in terms of computing resources coupled with the lack of connectivity constitute quite significant barriers.

%\section*{References}

\bibliography{BiblioSurvey}

@incollection{b1,
  author = {N. Indurkhya and F.J. Damerau },
  title = {An Overview of Modern Speech Recognition Xuedong Huang and Li Deng},
  edition = {0},
  publisher = {Chapman and Hall/CRC},
  year = 2010,
  pages = {363–390},
  doi = {10.1201/9781420085938-24.},
  language = {en},
  booktitle = {Handbook of Natural Language Processing}
}

@article{b2,
  citation-number = {2},
  author = {Benkerzaz , S. and Elmir Y. and Dennai A.},
  title = {A Study on Automatic Speech Recognition},
  volume = {10},
  pages = {10},
 year = {2019},
  issue = {3},
   journal = {Journal of  Information Technology Review },
  language = {en},
  number = {3}
}

@article{b3,
  citation-number = {3},
  author = {Yadava, T.G. and Jayanna, H.S.},
  title = {A spoken query system for the agricultural commodity prices and weather information access in Kannada language},
  volume = {20},
  pages = {635–644,},
 year = {2017},
  language = {en},
  journal = {Int. J. Speech Technol},
  number = {3}
}

@inproceedings{b6,
  citation-number = {6},
  author = {Husni, H. and Jamaluddin, Z.},
  title = {A retrospective and future look at speech recognition applications in assisting children with reading disabilities},
 year = {2008},
  language = {en},
  booktitle = {Proceedings of the world Congress on Engineering and Computer Science},
  address = {San Francisco, USA}
}

@inproceedings{b7,
  citation-number = {7},
  author = {Mayer, J.},
  title = {Low Cost Automatic Speech Recognition IOT Architecture for Hands Free Assistance to People with Mobility Restrictions},
  language = {en},
   year = {2018},
  pages = {53–58},
  booktitle = {Proceedings on the International Conference on Internet Computing (ICOMP)}
}

@article{b8,
  citation-number = {8},
  author = {Besacier, L. and Barnard, E. and Karpov, A. and Schultz, T.},
  title = {Automatic speech recognition for under-resourced languages: A survey},
  volume = {56},
  pages = {85–100,},
 year = {2014},
  doi = {10.1016/j.specom.2013.07.008.},
  language = {en},
  journal = {Speech Commun}
}

@article{b9,
  citation-number = {9},
  author = {Albabtain, A.F. and AlMulhim, D.A. and Yunus, F. and Househ, M.S.},
  title = {The role of mobile health in the developing world: a review of current knowledge and future trends},
  volume = {Informatics42},
  pages = {10–15,},
 year = {2014},
  language = {en},
  journal = {Cyber J. Multidiscip. J. Sci. Technol. JSHI J. Sel. Areas Health}
}

@incollection{b10,
  citation-number = {10},
  author = {Ebongue, J.L.F.K.},
  title = {Rethinking Network Connectivity in Rural Communities in Cameroon},
 year = {2015},
  note = {Online]. Available:},
  url = {https://arxiv.org/ftp/arxiv/paper/1505/1505.04449.pdf},
  language = {en},
  booktitle = {IST-Africa},
  address = {Lilongwe, Malaw}
}

@misc{b11,
  citation-number = {11},
  author = {Ebongue Louis, F.K.J.},
  title = {Wireless Mesh Network: a rural community case},
  publisher = {Universitat Bremen},
 year = {2015},
  note = {Accessed: Oct. 20, 2020. [Online]. Available:},
  url = {https://www.researchgate.net/publication/282297986_Wireless_Mesh_Network_a_rural_community_case},
  language = {fr}
}

@article{b12,
  citation-number = {12},
  author = {Saon, G. and Chien, J.-T.},
  title = {Large-vocabulary continuous speech recognition systems: A look at some recent advances},
  volume = {29},
  pages = {18–33,},
 year = {2012},
  language = {en},
  journal = {IEEE Signal Process. Mag},
  number = {6}
}

@article{b13,
  citation-number = {13},
  author = {Orosanu, L. and Jouvet, D.},
  title = {Adding New Words into a Language Model using Parameters of Known Words with Similar Behavior},
  volume = {128},
  pages = {18–24,},
 year = {2018},
  doi = {10.1016/j.procs.2018.03.003.},
  language = {en},
  journal = {Procedia Comput. Sci}
}

@misc{b14,
  citation-number = {14},
  author = {Warden, P.},
  title = {Speech Commands: A Dataset for Limited-Vocabulary Speech Recognition},
   year = {2018},
  note = {ArXiv180403209 Cs, Apr. 2018, Accessed: Jan. 02, 2020. [Online]. Available:},
  url = {http://arxiv.or},
  language = {en}
}

@article{b15,
  citation-number = {15},
  author = {Lima, T.Aguiar and Costa-Abreu, M.Da},
  title = {A survey on automatic speech recognition systems for Portuguese language and its variations},
  volume = {62},
  pages = {101055,},
 year = {2020},
  doi = {10.1016/j.csl.2019.101055.},
  language = {en},
  journal = {Comput. Speech Lang}
}

@article{b16,
  citation-number = {16},
  author = {Kurian, C.},
  title = {A Survey on Speech Recogntion in Indian Languages},
  volume = {5},
  pages = {6169-6175},
 year = {2014},
  language = {en},
  journal={/ (IJCSIT) International Journal of Computer Science and Information Technologies}, 
  volume={5}
}

@article{b17,
  citation-number = {17},
  author = {Ronzhin, A.L. and Yusupov, R.M. and Li, I.V. and Leontieva, A.B.},
  title = {Survey of Russian Speech Recognition Systems},
  pages = {7,},
 year = {2006},
  language = {en},
  journal = {St Petersburg}
}

@article{b18,
  citation-number = {18},
  author = {Minaee, S. and Abdolrashidi, A. and Su, H. and Bennamoun, M. and Zhang, D.},
  title = {Biometric Recognition Using Deep Learning: A Survey},
 year = {2020},
  volume = {12},
  note = {Online]. Available:},
  url = {http://arxiv.org/abs/1912.00271},
  language = {en},
  journal = {ArXiv191200271 Cs, Apr}
}

@article{b19,
  citation-number = {19},
  author = {Desai, N. and Dhameliya, K. and Desai, V.},
  title = {Feature Extraction and Classification Techniques for Speech Recognition: A Review},
  volume = {3},
  pages = {5,},
 year = {2013},
  issue = {12},
  journal = {International Journal of Emerging Technology and Advanced Engineering},
  language = {en}
}

@article{b20,
  citation-number = {20},
  author = {Hibare, R. and Vibhute, A.},
  title = {Feature Extraction Techniques in Speech Processing: A Survey},
  volume = {107},
  pages = {1–8,},
 year = {2014},
  doi = {10.5120/18744-9997.},
  language = {fr},
  journal = {Int. J. Comput. Appl},
  number = {5}
}

@article{b21,
  citation-number = {21},
  author = {Gong, Y.},
  title = {Speech recognition in noisy environments: A survey},
  volume = {16},
  pages = {261–291,},
 year = {1995},
  doi = {10.1016/01676393(94)00059-J.},
  language = {fr},
  journal = {Speech Commun},
  number = {3}
}

@article{b22,
  citation-number = {22},
  author = {Zaykovskiy, D.},
  title = {Survey of the Speech Recognition Techniques for Mobile Devices},
  pages = {6,},
 year = {2006},
  language = {en},
  journal = {St Petersburg}
}

@article{b23,
  citation-number = {23},
  author = {Sameera Abdul-Kader and Dr. John Woods},
  title = {Survey on Chatbot Design Techniques in Speech Conversation Systems},
  volume = {6},
 year = {2015},
  doi = {10.14569/IJACSA.2015.060712.},
  language = {fr},
  journal = {International Journal of Advanced Computer Science and Applications},
  number = {7}
}

@article{b24,
  citation-number = {24},
  author = {Jiang, H.},
  title = {Confidence measures for speech recognition: A survey},
  volume = {45},
  pages = {455–470,},
 year = {2005},
  doi = {10.1016/j.specom.2004.12.004.},
  language = {fr},
  journal = {Speech Commun},
  number = {4}
}

@misc{b25,
  citation-number = {25},
  author = {Agnes, M. and Guralnik, D.B.},
  title = {Webster’s new world college dictionary},
 year = {1999},
  language = {en}
}

@article{b26,
  citation-number = {26},
  author = {Clements, G.N.},
  title = {The geometry of phonological features},
  volume = {2},
  pages = {225–252,},
 year = {1985},
  language = {en},
  journal = {Phonology},
  number = {1}
}

@article{b27,
  citation-number = {27},
  author = {Saksamudre, S. and Shrishrimal, P.P. and Deshmukh, R.},
  title = {A Review on Different Approaches for Speech Recognition System},
  volume = {115},
  pages = {23–28,},
 year = {2015},
  doi = {10.5120/20284-2839.},
  language = {en},
  journal = {Int. J. Comput. Appl}
}

@inproceedings{b28,
  citation-number = {28},
  author = {Whittaker, E.W.D. and Woodland, R.C.},
  title = {Efficient class-based language modelling for very large vocabularies},
 year = {2001},
  volume = {1},
  pages = {545–548},
  doi = {10.1109/ICASSP.2001.940889},
  language = {en},
  booktitle = {2001 IEEE International Conference on Acoustics, Speech, and Signal Processing. Proceedings (Cat.No.01CH37221)},
  address = {Salt Lake City, UT, USA}
}

@article{b29,
  citation-number = {29},
  author = {Ghai, W. and Singh, N.},
  title = {Literature Review on Automatic Speech Recognition},
  volume = {41},
  pages = {42–50,},
 year = {2012},
  doi = {10.5120/5565-7646.},
  language = {en},
  journal = {Int. J. Comput. Appl},
  number = {8}
}

@article{b30,
  citation-number = {30},
  author = {Kraleva, R. and Kralev, V.},
  title = {On model architecture for a children’s speech recognition interactive dialog system},
  pages = {6,},
 year = {2009},
  language = {it},
  journal = {Nat. Sci}
}

@misc{b31,
  citation-number = {31},
  title = {Feature Extraction for ASR: Intro| Formula Coding},
  author = {Ravindra Lokhande},
  url = {http://wantee.github.io/2015/03/14/feature-extraction-for-asr-intro/},
  year = {2015},
  note = {accessed Jul. 01, 2021).},
  language = {en}
}

@inproceedings{b32,
  citation-number = {32},
  author = {Doukas, N. and Bardis, N.G. and Markovskyi, O.P.},
  title = {Task and Context Aware Isolated Word Recognition},
 year = {2017},
  pages = {330–334},
  language = {en},
  booktitle = {2017 International Conference on Control, Artificial Intelligence, Robotics and Optimization (ICCAIRO)}
}

@inproceedings{b33,
  citation-number = {33},
  author = {Doukas, N. and Bardis, N.G.},
  title = {Current trends in small vocabulary speech recognition for equipment control},
 year = {2017},
  pages = {020029},
  doi = {10.1063/1.4996686.},
  booktitle = {MATHEMATICAL METHODS AND COMPUTATIONAL TECHNIQUES IN SCIENCE AND ENGINEERING},
  language = {en},
  address = {Cambridge, UK}
}

@article{b34,
  citation-number = {34},
  author = {Narang, S. and Gupta, D.},
  title = {Speech Feature Extraction Techniques: A Review},
  pages = {107-114},
 year = {2015},
 volume={4},
  language = {en},
  journal={ International Journal of Computer Science and Mobile Computing},
  number={3}
}

@article{b35,
  citation-number = {35},
  author = {O’Shaughnessy, D.},
  title = {Linear predictive coding},
  volume = {7},
  pages = {29–32,},
 year = {1988},
  language = {en},
  journal = {IEEE Potentials},
  number = {1}
}

@article{b36,
  citation-number = {36},
  author = {Hermansky, H. and Morgan, N.},
  title = {RASTA processing of speech},
  volume = {2},
  pages = {578–589,},
 year = {1994},
  language = {en},
  journal = {IEEE Trans. Speech Audio Process},
  number = {4}
}

@inproceedings{b37,
  citation-number = {37},
  author = {Ioffe, S.},
  title = {Probabilistic linear discriminant analysis},
 year = {2006},
  pages = {531–542},
  language = {en},
  booktitle = {European Conference on Computer Vision}
}

@misc{b40,
  citation-number = {38},
  author = {Bluche, T. and Gisselbrecht, T.},
  title = {Predicting detection filters for small footprint open-vocabulary keyword spotting},
 year = {1912},
  note = {Accessed: Jan. 02, 2020. [Online]. Available:},
  url = {http://arxiv.org/abs/1912.07575},
  language = {en}
}

@article{b41,
  citation-number = {39},
  author = {Mittal, P. and Singh, N.},
  title = {Subword analysis of small vocabulary and large vocabulary ASR for Punjabi language},
  volume = {23},
  pages = {71–78,},
 year = {2020},
  doi = {10.1007/s10772-020-09673-3.},
  language = {en},
  journal = {Int. J. Speech Technol},
  number = {1}
}

@inproceedings{b42,
  citation-number = {40},
    author = {Gerazov, B. and Ivanovski, Z.},
  title = {A Speaker Independent Small Vocabulary Automatic Speech Recognition System in Macedonian},
  language = {en},
  booktitle = {Proceedings of the Second International Conference TAKTONS},
 year = {2013},
  pages = {5},
}

@article{b43,
  citation-number = {41},
  author = {Huang, F.L.},
  title = {An Effective Approach for Chinese Speech Recognition on Small size of Vocabulary},
  volume = {2},
  pages = {48–60,},
 year = {2011},
  doi = {10.5121/sipij.2011.2205.},
  language = {en},
  journal = {Signal Image Process. Int. J},
  number = {2}
}

@article{b44,
  citation-number = {42},
  author = {Liu, X.F. and Zhou, H.},
  title = {A Chinese Small Vocabulary Offline Speech Recognition System Based on Pocketsphinx in Android Platform},
  volume = {623},
  pages = {267–273,},
 year = {2014},
  doi = {10.4028/www.scientific.net/AMM.623.267.},
  language = {en},
  journal = {Appl. Mech. Mater}
}

@inproceedings{b45,
  citation-number = {43},
  author = {Fish, R.},
  title = {Using Audio Quality to Predict Word Error Rate in an Automatic Speech Recognition System},
  year = {2006},
  pages = {4},
  url={https://www.mitre.org/sites/default/files/pdf/06_1154.pdf},
  language = {en}
}

@book{b46,
  citation-number = {44},
  author = {Renals, S. and Grefenstette, G.},
  title = {Text-and speech-triggered information access: 8th ELSNET Summer School, Chios Island},
 year = {2000},
  edition = {revised lectures},
  publisher = {Springer},
  language = {da},
  address = {Greece}
}

@inproceedings{b47,
  citation-number = {45},
  author = {Tamgno, J.K. and Barnard, E. and Lishou, C. and Richomme, M.},
  title = {Wolof Speech Recognition Model of Digits and Limited-Vocabulary Based on HMM and ToolKit},
  language = {en},
  booktitle = {2012 UKSim 14th International Conference on Computer Modelling and Simulation, Cambridge United Kingdom},
 year = {2012},
  pages = {389–395},
  doi = {10.1109/UKSim.2012.118.}
}

@article{b49,
  citation-number = {47},
  author = {Baggenstoss, P.M.},
  title = {A modified Baum-Welch algorithm for hidden Markov models with multiple observation spaces},
  volume = {9},
  pages = {411–416,},
 year = {2001},
  language = {en},
  journal = {IEEE Trans. Speech Audio Process},
  number = {4}
}

@article{b50,
  citation-number = {48},
  author = {Yu, S.-Z. and Kobayashi, H.},
  title = {An efficient forward-backward algorithm for an explicit-duration hidden Markov model},
  volume = {10},
  pages = {11–14,},
 year = {2003},
  language = {en},
  journal = {IEEE Signal Process. Lett},
  number = {1}
}

@inproceedings{b51,
  citation-number = {49},
  author = {Militaru, D. and Lazar, M.},
  title = {LIMITED-VOCABULARY SPEAKER INDEPENDENT CONTINUOUS SPEECH RECOGNITION USING CLASSBASED LANGUAGE MODEL},
  volume = {58},
  pages = {6},
 year = {2017},
  issue = {1},
  booktitle={Proceedings of the Annual Symposium of the Institute of Solid Mechanics
 and Session of the Commission of Acoustics},
  language = {vi}
}

@misc{b52,
  citation-number = {50},
  author = {Fosler-Lussier, J.E.},
  title = {Dynamic Pronunciation Models for Automatic Speech Recognition},
  pages = {204},
  year = {1999},
  note = {INTERNATIONAL COMPUTER SCIENCE INSTITUTE},
  url={https://citeseerx.ist.psu.edu/viewdoc/download?doi=10.1.1.42.7647&rep=rep1&type=pdf},
  language = {en}
}

@incollection{b53,
  citation-number = {51},
  author = {Fosler-Lussier, E.},
  title = {A Tutorial on Pronunciation Modeling for Large Vocabulary Speech Recognition},
  volume = {2705},
  editor = {Renals, S. and Grefenstette, G.},
  publisher = {Springer},
 year = {2003},
  pages = {38–77},
  doi = {10.1007/978-3-54045115-0_3.},
  language = {en},
  booktitle = {Text- and Speech-Triggered Information Access},
  address = {Berlin, Heidelberg}
}

@article{b54,
  citation-number = {52},
  author = {Viterbi, A.},
  title = {Error bounds for convolutional codes and an asymptotically optimum decoding algorithm},
  volume = {13},
  pages = {260–269,},
 year = {1967},
  language = {en},
  journal = {IEEE Trans. Inf. Theory},
  number = {2}
}

@article{b55,
  citation-number = {53},
  author = {Hui, J.},
  title = {Speech Recognition — ASR Decoding},
 year = {2019},
  url = {https://jonathan-hui.medium.com/speech-recognition-asr-decodingf152aebed779},
  note = {accessed Jun. 09, 2021).},
  language = {en},
  journal = {Medium}
}

@incollection{b56,
  citation-number = {54},
  author = {Novak, M.},
  title = {Evolution of the ASR Decoder Design},
  volume = {6231},
  editor = {Sojka, P. and Horak, A. and Kopecek, I. and Pala, K.},
  publisher = {Springer},
 year = {2010},
  pages = {10–17},
  doi = {10.1007/978-3-642-15760-8_3.},
  language = {nl},
  booktitle = {Text, Speech and Dialogue},
  address = {Berlin, Heidelberg}
}

@article{b57,
  citation-number = {55},
 author = { Erin Myers},
  title = {The Role of Artificial Intelligence and Machine Learning in Speech Recognition},
 year = {2019},
  url = {https://www.rev.com/blog/artificialintelligence-machine-learning-speech-recognition},
  note = {accessed Feb. 06, 2020).},
  language = {en},
  journal = {Rev}
}

@article{b58,
  citation-number = {56},
  author = {Murshed, M.G.S. and Murphy, C. and Hou, D. and Khan, N. and Ananthanarayanan, G. and Hussain, F.},
  title = {Machine Learning at the Network Edge: A Survey},
 year = {2020},
  volume = {15},
  note = {Online]. Available:},
  url = {http://arxiv.org/abs/1908.00080},
  language = {en},
  journal = {ArXiv190800080 Cs Stat}
}

@book{b59,
  citation-number = {57},
  author = {Murphy, K.P.},
  title = {Machine learning: a probabilistic perspective},
  publisher = {MIT Press},
 year = {2012},
  language = {en},
  address = {Cambridge, MA}
}

@article{b60,
  citation-number = {58},
  author = {Padmanabhan, J. and Premkumar, M.J.Johnson},
  title = {Machine Learning in Automatic Speech Recognition: A Survey},
  volume = {32},
  pages = {240–251,},
 year = {2015},
  doi = {10.1080/02564602.2015.1010611.},
  language = {en},
  journal = {IETE Tech. Rev},
  number = {4}
}

@incollection{b61,
  citation-number = {59},
  author = {Li Jinyu and Deng Li and Haeb-Umbach Reinhold and Gong Yifan},
  title = {Robust Automatic Speech Recognition},
  publisher = {Elsevier},
 year = {2016},
  pages = {1–7},
  language = {en}
}

@ARTICLE{b62,

  author={Hinton, Geoffrey and Deng, Li and Yu, Dong and Dahl, George E. and Mohamed, Abdel-rahman and Jaitly, Navdeep and Senior},

  journal={IEEE Signal Processing Magazine}, 
  title={Deep Neural Networks for Acoustic Modeling in Speech Recognition: The Shared Views of Four Research Groups}, 
  year={2012},
  volume={29},
  number={6},
  pages={82-97},
  doi={10.1109/MSP.2012.2205597}}

@inproceedings{b63,
  citation-number = {61},
  author = {Dhankar Abhishek},
  title = {Study of deep learning and CMU sphinx in automatic speech recognition},
 year = {2017},
  pages = {2296–2301},
   booktitle = {International Conference on Advances in Computing, Communications and Informatics (ICACCI)},
  address = {Udupi},
  language = {en}
 
}

@article{b64,
  citation-number = {62},
  author = {Hernández-Blanco, A. and Herrera-Flores, B. and Tomás, D. and NavarroColorado, B.},
  title = {A Systematic Review of Deep Learning Approaches to Educational Data Mining},
  volume = {2019},
  pages = {1–22,},
 year = {2019},
  doi = {doi: 10.1155/2019/1306039.},
  language = {en},
  journal = {Complexity}
}

@inproceedings{b65,
  citation-number = {63},
  author = {Dhankar, A.},
  title = {Study of deep learning and CMU sphinx in automatic speech recognition},
 year = {2017},
  pages = {2296–2301},
  doi = {10.1109/ICACCI.2017.8126189.},
  language = {en},
  booktitle = {2017 International Conference on Advances in Computing, Communications and Informatics (ICACCI)},
  address = {Udupi}
}

@article{b66,
  citation-number = {64},
  author = {Ko, T. and Peddinti, V. and Povey, D. and Khudanpur, S.},
  title = {Audio augmentation for speech recognition},
 year = {2015},
 journal={interspeech},
  language = {en}
}

@article{b67,
  citation-number = {65},
  author = {Salamon, J. and Bello, J.P.},
  title = {Deep convolutional neural networks and data augmentation for environmental sound classification},
  volume = {24},
  pages = {279–283,},
 year = {2017},
  language = {en},
  journal = {IEEE Signal Process. Lett},
  number = {3}
}

@misc{b68,
  citation-number = {66},
  author = {Park, D.S.},
  title = {Specaugment: A simple data augmentation method for automatic speech recognition},
  note = {ArXiv Prepr. ArXiv190408779,},
 year = {2019},
  language = {en}
}

@misc{b69,
  citation-number = {67},
  author = {Maguolo, G. and Paci, M. and Nanni, L. and Bonan, L.},
  title = {Audiogmenter: A MATLAB toolbox for audio data augmentation},
  note = {ArXiv Prepr. ArXiv191205472,},
 year = {2019},
  language = {fr}
}

@misc{b70,
  citation-number = {68},
   author = {Volodymyr Mnih, Koray Kavukcuoglu, David Silver},
  title = {deep reinforcement learning projects},
  year = {2020},
  url = {http://www.polodepensamento.com.br/blog/article.php?99f49d=deepreinforcement-learning-projects},
  note = {accessed Aug. 27, 2020},
  language = {en}
}

@misc{b71,
  citation-number = {69},
  author = {Rajapakshe, T. and Latif, S. and Rana, R. and Khalifa, S. and Schuller, B.W.},
  title = {Deep Reinforcement Learning with Pre-training for Time-efficient Training of Automatic Speech Recognition},
  note = {ArXiv Prepr. ArXiv200511172,},
 year = {2020},
  language = {en}
}

@inproceedings{b72,
  citation-number = {70},
  author = {Kala, T. and Shinozaki, T.},
  title = {Reinforcement learning of speech recognition system based on policy gradient and hypothesis selection},
 year = {2018},
  pages = {5759–5763},
    booktitle = {2018 IEEE International Conference on Acoustics, Speech and Signal Processing (ICASSP)},
  language = {en}
}

@article{b73,
  citation-number = {71},
  author = {Baker, J.},
  title = {Developments and directions in speech recognition and understanding, Part 1 [DSP Education]},
  volume = {26},
  pages = {75–80,},
 year = {2009},
  doi = {10.1109/MSP.2009.932166.},
  language = {en},
  journal = {IEEE Signal Process. Mag},
  number = {3}
}

@incollection{b74,
  citation-number = {72},
  author = {McDonough, J. and Wolfel, M.},
  title = {Distant Speech Recognition: Bridging the Gaps},
 year = {2008},
  pages = {108–114},
  doi = {10.1109/HSCMA.2008.4538699.},
  language = {en},
  booktitle = {2008 Hands-Free Speech Communication and Microphone Arrays},
  address = {Trento, Italy}
}

@book{b75,
  citation-number = {73},
  author = {Cohen, I. and Benesty, J. and Gannot, S.},
  title = {Speech processing in modern communication: challenges and perspectives},
  publisher = {Springer},
 year = {2010},
  language = {en},
  address = {Berlin}
}

@book{b76,
  citation-number = {74},
  author = {Virtanen, T. and Singh, R. and Raj, B.},
  title = {Techniques for Noise Robustness in Automatic Speech Recognition: Virtanen/Techniques for Noise Robustness in Automatic Speech Recognition},
  publisher = {John Wiley and Sons, Ltd},
 year = {2012},
  doi = {10.1002/9781118392683.},
  language = {fr},
  address = {Chichester, UK}
}

@incollection{b77,
  citation-number = {75},
  author = {Li, J. and Deng, L. and Haeb-Umbach, R. and Gong, Y.},
  title = {Fundamentals of speech recognition},
  publisher = {Elsevier},
 year = {2016},
  pages = {9–40},
  doi = {10.1016/B978-0-12-802398-3.00002-7.},
  language = {it},
  booktitle = {Robust Automatic Speech Recognition}
}

@book{b78,
  citation-number = {76},
  author = {Vincent, E. and Virtanen, T. and Gannot, S.},
  title = {Audio source separation and speech enhancement},
  publisher = {John Wiley and Sons},
 year = {2018},
  language = {fr},
  address = {Hoboken, NJ}
}

@article{b79,
  citation-number = {77},
  author = {Glasser, A.},
  title = {Automatic Speech Recognition Services: Deaf and Hard-of Hearing Usability},
 year = {2019},
  pages = {1–6,},
  doi = {10.1145/3290607.3308461.},
  language = {en},
  journal = {Ext. Abstr}
}

@article{b80,
  citation-number = {78},
  author = {Hofe, R.},
  title = {Small-vocabulary speech recognition using a silent speech interface based on magnetic sensing},
  volume = {55},
  pages = {22–32,},
 year = {2013},
  doi = {10.1016/j.specom.2012.02.001.},
  language = {fr},
  journal = {Speech Commun},
  number = {1}
}

@article{b81,
  citation-number = {79},
  author = {Alumae, T. and Vohandu, L.},
  title = {Limited-Vocabulary Estonian Continuous Speech Recognition System using Hidden Markov Models},
  year = {2004},
  volume = {15},
  pages = {303-314},
  journal = {Informatica, Institute of Mathematics and Informatics},
  number = {3},
  language = {en}
}

@misc{b82,
  citation-number = {80},
  author = {Zhou, K. and Sisman, B. and Liu, R. and Li, H.},
  title = {Emotional Voice Conversion: Theory, Databases and ESD},
  note = {ArXiv Prepr. ArXiv210514762, 2021.},
  year = {2021},
  language = {en}
}

@article{b83,
  citation-number = {81},
  author = {Bertin, N.},
  title = {VoiceHome-2, an extended corpus for multichannel speech processing in real homes},
  year = {2021},
  volume = {106},
  pages = {68–78,},
 year = {2019},
  doi = {10.1016/j.specom.2018.11.002.},
  language = {en},
  journal = {Speech Commun}
}

@article{b84,
  citation-number = {82},
  author = {Mezzoudj, F. and Langlois, D. and Jouvet, D. and Benyettou, A.},
  title = {Textual Data Selection for Language Modelling in the Scope of Automatic Speech Recognition},
  volume = {128},
  pages = {55–64,},
 year = {2018},
  doi = {10.1016/j.procs.2018.03.008.},
  language = {en},
  journal = {Procedia Comput. Sci}
}

@inproceedings{b85,
  citation-number = {83},
  author = {Pleshkova, S. and Bekyarski, A. and Zahariev, Z.},
  title = {Reduced Database for Voice Commands Recognition Using Cloud Technologies, Artificial Intelligence and Deep Learning},
 year = {2019},
  pages = {1–4},
  doi = {10.1109/ELMA.2019.8771526.},
  language = {en},
  booktitle = {2019 16th Conference on Electrical Machines, Drives and Power Systems (ELMA)},
  address = {Varna, Bulgaria}
}

@inproceedings{b86,
  citation-number = {84},
  author = {Petridis, S. and Shen, J. and Cetin, D. and Pantic, M.},
  title = {Visual-only recognition of normal, whispered and silent speech},
 year = {2018},
  pages = {6219–6223},
  language = {en},
  booktitle = {2018 IEEE International Conference on Acoustics, Speech and Signal Processing (ICASSP)}
}

@misc{b87,
  citation-number = {85},
  author = {Asuncion, A. and Newman, D.},
  title = {UCI machine learning repository},
 year = {2007},
  language = {pt},
  address = {Irvine, CA, USA}
}

@incollection{b88,
  citation-number = {86},
  author = {Qader, A. and Hassani, H.},
  title = {Kurdish (Sorani) Speech to Text: Presenting an Experimental Dataset},
 year = {2019},
  note = {Accessed: Mar. 26, 2020. [Online]. Available:},
  url = {http://arxiv.org/abs/1911.13087},
  language = {it},
  booktitle = {ArXiv191113087 Cs Eess}
}

@misc{b89,
      title={Common Voice: A Massively-Multilingual Speech Corpus}, 
      author={Rosana Ardila and Megan Branson and Kelly Davis and Michael Henretty and Michael Kohler and Josh Meyer},
      year={2020},
      note = {ArXiv Prepr. ArXiv191206670,},
      eprint={1912.06670},
      archivePrefix={arXiv},
      language = {en},
      primaryClass={cs.CL}
}

@inproceedings{b90,
  citation-number = {88},
  author = {Johnson, S.E. and Jourlin, P. and Moore, G.L. and Jones, K.S. and Woodland, P.C.},
  title = {The Cambridge University spoken document retrieval system},
 year = {1999},
  pages = {49–52},
  volume = {1},
  doi = {10.1109/ICASSP.1999.758059.},
  language = {en},
  series = {Cat. No.99CH36258},
  booktitle = {1999 IEEE International Conference on Acoustics, Speech, and Signal Processing. Proceedings. ICASSP99},
  address = {Phoenix, AZ, USA}
}

@article{b91,
  citation-number = {89},
  author = {Anh, N.T. and Thi, H.},
  title = {KWA: A New Method of Calculation and Representation Accuracy for Speech Keyword Spotting in String Results},
  volume = {11},
 year = {2020},
  doi = {10.14569/IJACSA.2020.0110283.},
  language = {en},
  journal = {International Journal of Advanced Computer Science and Applications},
  number = {2}
}

@inproceedings{b92,
  citation-number = {90},
  author = {Kurimo, M.},
  title = {Unlimited vocabulary speech recognition for agglutinative languages},
 year = {2006},
  pages = {487–494},
  language = {en},
  booktitle = {Proceedings of the Human Language Technology Conference of the NAACL, Main Conference}
}

@misc{b93,
  citation-number = {91},
  author = {Kurimo, M. and Creutz, M. and Varjokallio, M. and Arsoy, E. and Saraclar, M.},
  title = {Unsupervised segmentation of words into morphemes-Morpho challenge 2005 Application to automatic speech recognition},
 year = {2006},
  language = {en}
}

@misc{b94,
  citation-number = {92},
  author = {Huang, C. and Chang, E. and Zhou, J. and Lee, K.-F.},
  title = {Accent modeling based on pronunciation dictionary adaptation for large vocabulary Mandarin speech recognition},
 year = {2000},
  language = {en}
}

@incollection{b95,
  citation-number = {93},
  author = {Galibert, O.},
  title = {Methodologies for the evaluation of speaker diarization and automatic speech recognition in the presence of overlapping speech},
 year = {2013},
  pages = {1131–1134},
  language = {en},
  booktitle = {INTERSPEECH}
}

@article{b96,
  citation-number = {94},
  author = {Jelinek, F. and Mercer, R.L. and Bahl, L.R. and Baker, J.K.},
  title = {Perplexity—a measure of the difficulty of speech recognition tasks},
  volume = {62},
  pages = {63– 63,},
 year = {1977},
  doi = {10.1121/1.2016299.},
  language = {en},
  journal = {J. Acoust. Soc. Am},
  number = {S1}
}

@article{b97,
  citation-number = {95},
  author = {Shannon, C.E.},
  title = {Prediction and entropy of printed English},
  volume = {30},
  pages = {50–64,},
 year = {1951},
  language = {en},
  journal = {Bell Syst. Tech. J},
  number = {1}
}

@article{b98,
  citation-number = {96},
  author = {Matarneh, R. and Maksymova, S. and Lyashenko, V.V. and Belova, N.V.},
  title = {Speech Recognition Systems: A Comparative Review},
  volume = {19},
  year = {2017},
  pages = {71-79},
  language = {en},
  journal = {Journal of Computer Engineering (IOSR-JCE)},
  number = {5}
}

@inproceedings{b99,
  citation-number = {97},
  author = {Fujiwara, K.},
  title = {Error correction of speech recognition by custom phonetic alphabet input for ultra-small devices},
 year = {2016},
  pages = {104–109},
  language = {en},
  booktitle = {Proceedings of the 2016 CHI Conference Extended Abstracts on Human Factors in Computing Systems}
}

@inproceedings{b100,
  citation-number = {98},
  author = {Prozorov, D. and Tatarinova, A.},
  title = {Comparison Of Grapheme-to-Phoneme Conversions For Spoken Document Retrieval},
 year = {2019},
  pages = {1–4},
  language = {en},
  booktitle = {2019 IEEE East-West Design and Test Symposium (EWDTS)}
}

@incollection{b101,
  citation-number = {99},
  author = {Rawat, S. and Gupta, P. and Kumar, P.},
  title = {Digital life assistant using automated speech recognition},
 year = {2014},
  pages = {43–47},
  doi = {10.1109/CIPECH.2014.7019075.},
  language = {en},
  booktitle = {2014 Innovative Applications of Computational Intelligence on Power, Energy and Controls with their impact on Humanity (CIPECH)},
  address = {Ghaziabad, UP, India}
}

@article{b102,
  citation-number = {100},
  author = {Dietterich, T.G. and Horvitz, E.J.},
  title = {Rise of concerns about AI: reflections and directions},
  volume = {58},
  pages = {38–40,},
 year = {2015},
  doi = {10.1145/2770869.},
  language = {fr},
  journal = {Commun. ACM},
  number = {10}
}

@article{b103,
  citation-number = {101},
  author = {Zeng, D.},
  title = {AI Ethics: Science Fiction Meets Technological Reality},
  volume = {30},
  pages = {2–5,},
 year = {2015},
  doi = {10.1109/MIS.2015.53.},
  language = {en},
  journal = {IEEE Intell. Syst},
  number = {3}
}

@inproceedings{b104,
  citation-number = {102},
  author = {Shibata, D. and Wakamiya, S. and Ito, K. and Miyabe, M. and Kinoshita, A. and Aramaki, E.},
  title = {VocabChecker: measuring language abilities for detecting early stage dementia},
 year = {2018},
  pages = {1–2},
  language = {en},
  booktitle = {Proceedings of the 23rd International Conference on Intelligent User Interfaces Companion}
}

@incollection{b105,
  citation-number = {103},
  author = {Herbert, D. and Kang, B.},
  title = {Comparative analysis of intelligent personal agent performance},
 year = {2019},
  pages = {127–141},
  language = {en},
  booktitle = {Pacific Rim Knowledge Acquisition Workshop}
}

@misc{b106,
  citation-number = {104},
  author = {Gaida, C. and Lange, P. and Petrick, R. and Proba, P. and Malatawy, A. and Suendermann-Oeft, D.},
  title = {Comparing open-source speech recognition toolkits},
 year = {2014},
 url={http://antikenschlacht.de/su/pdf/oasis2014.pdf},
  language = {en}
}

@article{b107,
  citation-number = {105},
  author = {Narayanaswami, A. and Vassiliadis, C.},
  title = {An online prototype speechenabled information access tool using Java Speech Application Programming Interface},
  volume = {},
 year = {2001},
  pages = {111–115},
  language = {en},
  journal = {Proceedings of the 33rd Southeastern Symposium on System Theory (Cat)},
  number = {01EX460)}
}

@article{b108,
  citation-number = {106},
  author = {Shi, H. and Maier, A.},
  title = {Speech enabled shopping application using Microsoft SAPI},
  volume = {6},
  pages = {33,},
 year = {1996},
  language = {en},
  journal = {Int. J. Comput. Sci. Netw. Secur},
  number = {9}
}

@article{b109,
  citation-number = {107},
  author = {Këpuska, V.},
  title = {Comparing Speech Recognition Systems (Microsoft API, Google API And CMU Sphinx)},
  volume = {07},
  pages = {20–24,},
 year = {2017},
  doi = {10.9790/9622-0703022024.},
  language = {en},
  journal = {Int. J. Eng. Res. Appl},
  number = {03}
}

@article{b110,
  citation-number = {108},
  author = {Young, S.},
  title = {The HTK book},
  volume = {3},
  pages = {12,},
 year = {2002},
  language = {en},
  journal = {Camb. Univ. Eng. Dep},
  number = {175}
}

@inproceedings{b111,
  citation-number = {109},
  author = {Supriya, S. and Handore, S.M.},
  title = {Speech recognition using HTK toolkit for Marathi language},
 year = {2017},
  pages = {1591–1597},
  doi = {10.1109/ICPCSI.2017.8391979.},
  language = {en},
  booktitle = {2017 IEEE International Conference on Power, Control, Signals and Instrumentation Engineering (ICPCSI)},
  address = {Chennai}
}

@inproceedings{b112,
  citation-number = {110},
  author = {Qiao, F. and Sherwani, J. and Rosenfeld, R.},
  title = {Small-vocabulary speech recognition for resource-scarce languages},
 year = {2010},
  pages = {1},
  doi = {10.1145/1926180.1926184.},
  language = {en},
  booktitle = {Proceedings of the First ACM Symposium on Computing for Development - ACM DEV ’10},
  address = {London, United Kingdom}
}

@misc{b113,
  citation-number = {111},
  author = {Hatala, Z.},
  title = {Practical speech recognition with htk},
  note = {ArXiv Prepr. ArXiv190802119,},
 year = {2019},
  language = {en}
}

@incollection{b114,
  citation-number = {112},
  author = {Veiga, A. and Lopes, C. and Sá, L. and Perdigão, F.},
  title = {Acoustic Similarity Scores for Keyword Spotting},
  volume = {8775},
  editor = {Baptista, J. and Mamede, N. and Candeias, S. and Paraboni, I. and Pardo, T.A.S. and G. Volpe Nunes, M.},
  publisher = {Springer International Publishing},
 year = {2014},
  pages = {48–58},
  doi = {10.1007/978-3-319-09761-9_5.},
  language = {en},
  booktitle = {Computational Processing of the Portuguese Language},
  address = {Cham}
}

@misc{b115,
  citation-number = {113},
  author = {Veras, J.C. and M. Prego, T. and Lima, A.A. and Ferreira, T.N. and Netto, S.L.},
  title = {Speech quality enhancement based on spectral subtraction},
 year = {2014},
  language = {en}
}

@misc{b116,
  citation-number = {114},
  author = {Zen, H. and Agiomyrgiannakis, Y. and Egberts, N. and Henderson, F. and Szczepaniak, P.},
  title = {Fast, Compact, and High Quality LSTM-RNN Based Statistical Parametric Speech Synthesizers for Mobile Devices},
  year = {2016},
  note = {ArXiv160606061 Cs, Jun. 2016, Accessed: Apr. 14, 2020. [Online]. Available:},
  url = {http://arxiv.org/abs/1606.06061},
  language = {en}
}

@book{b117,
  citation-number = {115},
  author = {Lee, K.-F.},
  title = {Automatic speech recognition: the development of the SPHINX system},
  volume = {62},
  publisher = {Springer Science and Business Media},
 year = {1988},
  language = {en}
}

@inproceedings{b118,
  citation-number = {116},
  author = {Huggins-Daines, D. and Kumar, M. and Chan, A. and Black, A.W. and Ravishankar, M. and Rudnicky, A.I.},
  title = {Pocketsphinx: A free, real-time continuous speech recognition system for hand-held devices},
 year = {2006},
  volume = {1},
  language = {en},
  booktitle = {2006 IEEE International Conference on Acoustics Speech and Signal Processing Proceedings}
}

@INPROCEEDINGS{b119,

  author={Schultz, Tanja},
  booktitle={2009 IEEE Workshop on Automatic Speech Recognition   Understanding}, 
  title={Rapid language adaptation tools for multilingual speech processing}, 
  year={2009},
  pages={51-51},
  doi={10.1109/ASRU.2009.5373503}}

@inproceedings{b120,
  citation-number = {118},
  author = {Schlippe, T. and Volovyk, M. and Yurchenko, K. and Schultz, T.},
  title = {Rapid bootstrapping of a ukrainian large vocabulary continuous speech recognition system},
 year = {2013},
  pages = {7329–7333},
  language = {en},
  booktitle = {2013 IEEE International Conference on Acoustics, Speech and Signal Processing}
}

@misc{b121,
  citation-number = {119},
  author = {Vu, N.T. and Schlippe, T. and Kraus, F. and Schultz, T.},
  title = {Rapid bootstrapping of five eastern european languages using the rapid language adaptation toolkit},
 year = {2010},
  language = {en}
}

@article{b122,
  citation-number = {120},
  author = {Povey, D.},
  title = {The Kaldi speech recognition toolkit},
 year = {2011},
  volume = {},
  publisher = {CONF},
  language = {en},
  journal = {IEEE 2011 workshop on automatic speech recognition and understanding},
  number = {}
}

@inproceedings{b123,
  citation-number = {121},
  author = {Cosi, P.},
  title = {A kaldi-dnn-based asr system for italian},
 year = {2015},
  pages = {1–5},
  language = {en},
  booktitle = {2015 international joint conference on neural networks (IJCNN)}
}

@article{b124,
  citation-number = {122},
  author = {Guglani, J. and Mishra, A.N.},
  title = {Continuous Punjabi speech recognition model based on Kaldi ASR toolkit},
  volume = {21},
  pages = {211–216,},
 year = {2018},
  language = {en},
  journal = {Int. J. Speech Technol},
  number = {2}
}

@inproceedings{b125,
  citation-number = {123},
  author = {Upadhyaya, P. and Farooq, O. and Abidi, M.R. and Varshney, Y.V.},
  title = {Continuous Hindi speech recognition model based on Kaldi ASR toolkit},
 year = {2017},
  pages = {786–789},
  language = {en},
  booktitle = {2017 International Conference on Wireless Communications, Signal Processing and Networking (WiSPNET)}
}

@inproceedings{b126,
  citation-number = {124},
  author = {Banerjee, D.S. and Hamidouche, K. and Panda, D.K.},
  title = {Re-designing CNTK deep learning framework on modern GPU enabled clusters},
 year = {2016},
  pages = {144–151},
  language = {en},
  booktitle = {2016 IEEE international conference on cloud computing technology and science (CloudCom)}
}

@inproceedings{b127,
  citation-number = {125},
  author = {Seide, F. and Agarwal, A.},
  title = {CNTK: Microsoft’s open-source deep-learning toolkit},
 year = {2016},
  pages = {2135–2135},
  language = {en},
  booktitle = {Proceedings of the 22nd ACM SIGKDD International Conference on Knowledge Discovery and Data Mining}
}

@misc{b128,
  citation-number = {126},
  author = {Paci, G. and Sommavilla, G. and Tesser, F. and Cosi, P.},
  title = {Julius ASR for Italian children speech},
 year = {2013},
 url={https://www.istc.cnr.it/en/content/julius-asr-italian-children-speech},
  language = {en}
}

@inproceedings{b129,
  citation-number = {127},
  author = {Sharma, R.S. and Paladugu, S.H. and Priya, K.J. and Gupta, D.},
  title = {Speech recognition in Kannada using HTK and julius: A comparative study},
 year = {2019},
  pages = {0068–0072},
  language = {en},
  booktitle = {2019 International Conference on Communication and Signal Processing (ICCSP)}
}

@inproceedings{b130,
  citation-number = {128},
  author = {Pleva, M. and Juhár, J. and Thiessen, A.S.},
  title = {Automatic Acoustic Speech segmentation in Praat using cloud based ASR},
 year = {2015},
  pages = {172–175},
  language = {en},
  booktitle = {2015 25th International Conference Radioelektronika (RADIOELEKTRONIKA)}
}

@inproceedings{b131,
  citation-number = {129},
  author = {Hateva, N. and Mitankin, P. and Mihov, S.},
  title = {BulPhonC: bulgarian speech corpus for the development of ASR technology},
 year = {2016},
  pages = {771–774},
  language = {en},
  booktitle = {Proceedings of the Tenth International Conference on Language Resources and Evaluation (LREC’16)}
}

@article{b132,
  citation-number = {130},
  author = {Hannun, A.},
  title = {Deep speech: Scaling up end-to-end speech recognition},
 year = {2014},
  language = {en},
  journal = {ArXiv Prepr. ArXiv14125567}
}

@article{b133,
  citation-number = {131},
  author = {Li, J.},
  title = {Jasper: An end-to-end convolutional neural acoustic model},
 year = {2019},
  language = {fr},
  journal = {ArXiv Prepr. ArXiv190403288}
}

@inproceedings{b134,
  citation-number = {132},
  author = {Afouras, T. and Chung, J.S. and Zisserman, A.},
  title = {ASR is all you need: Cross-modal distillation for lip reading},
 year = {2020},
  pages = {2143–2147},
  language = {en},
  booktitle = {ICASSP 2020-2020 IEEE International Conference on Acoustics, Speech and Signal Processing (ICASSP)}
}

@inproceedings{b135,
  citation-number = {133},
  author = {Liu, S. and Cao, Y. and Hu, N. and Su, D. and Meng, H.},
  title = {Fastsvc: Fast Cross-Domain Singing Voice Conversion With Feature-Wise Linear Modulation},
 year = {2021},
  pages = {1–6},
  language = {en},
  booktitle = {2021 IEEE International Conference on Multimedia and Expo (ICME)}
}

@misc{b136,
  citation-number = {134},
  author = {Liu, Y. and Iyer, R. and Kirchhoff, K. and Bilmes, J.},
  title = {SVitchboard II and FiSVer I: High-quality limited-complexity corpora of conversational English speech},
 year = {2015},
  language = {en}
}

@inproceedings{b137,
  citation-number = {135},
  author = {Lucey, S. and 
  ran, S. and Chandran, V.},
  title = {Improving visual noise insensitivity in small vocabulary audio visual speech recognition applications},
 year = {2001},
  volume = {2},
  pages = {434–437},
  doi = {10.1109/ISSPA.2001.950173.},
  language = {en},
  booktitle = {Proceedings of the Sixth International Symposium on Signal Processing and its Applications (Cat.No.01EX467), Kuala Lumpur},
  address = {Malaysia}
}

@article{b138,
  citation-number = {136},
  author = {Chaudhuri, S. and Raj, B. and Ezzat, T.},
  title = {A Paradigm for Limited Vocabulary Speech Recognition Based on Redundant Spectro-Temporal Feature Sets},
  year = {2011},
  pages = {4},
  journal={INTERSPEECH},
  language = {en}
}

@thesis{b139,
  citation-number = {137},
  author = {Haflan, V.},
  title = {Noise Robustness in Small Vocabulary Speech Recognition},
  year = {2019},
  language = {en},
  school = {NTNU},
  note = {Master's Thesis},
  url={http://hdl.handle.net/11250/2613396}
}

@inproceedings{b140,
  citation-number = {138},
  author = {Burileanu, D. and Sima, M. and Negrescu, C. and Croitoru, V.},
  title = {Robust recognition of small vocabulary telephone quality speech},
 year = {2003},
  language = {en},
   booktitle = {Proceedings of the Second Conference on Speech Technology and Human-Computer Dialogue (SpeD 2003)
Bucharest},
address = {Romania}
}

@inproceedings{b141,
  citation-number = {139},
  author = {Bali, K. and Sitaram, S. and Cuendet, S. and Medhi, I.},
  title = {A Hindi speech recognizer for an agricultural video search application},
 year = {2013},
  pages = {1},
  doi = {10.1145/2442882.2442889.},
  language = {en},
  booktitle = {Proceedings of the 3rd ACM Symposium on Computing for Development - ACM DEV ’13},
  address = {Bangalore, India}
}

@inproceedings{b142,
  citation-number = {140},
  author = {Al-Qatab, B.A.Q. and Ainon, R.N.},
  title = {Arabic speech recognition using Hidden Markov Model Toolkit(HTK)},
 year = {2010},
  pages = {557–562},
  doi = {10.1109/ITSIM.2010.5561391.},
  language = {en},
  booktitle = {2010 International Symposium on Information Technology, Kuala Lumpur},
  address = {Malaysia}
}

@inproceedings{b143,
  citation-number = {141},
  author = {Azmi, M.M. and Tolba, H.},
  title = {Syllable-based automatic Arabic speech recognition in different conditions of noise},
 year = {2008},
  pages = {601–604},
  doi = {10.1109/ICOSP.2008.4697204.},
  language = {en},
  booktitle = {2008 9th International Conference on Signal Processing},
  address = {Beijing, China}
}

@inproceedings{b144,
  citation-number = {142},
  author = {Muhammad, G. and Alotaibi, Y.A. and Huda, M.N.},
  title = {Automatic speech recognition for Bangla digits},
 year = {2009},
  pages = {379–383},
  doi = {10.1109/ICCIT.2009.5407267.},
  language = {en},
  booktitle = {2009 12th International Conference on Computers and Information Technology},
  address = {Dhaka, Bangladesh}
}

@article{b145,
  citation-number = {143},
  author = {Krishnan, V.R.V. and Jayakumar, A. and Babu, A.P.},
  title = {Speech Recognition of Isolated Malayalam Words Using Wavelet Features and Artificial Neural Network},
  publisher = {delta},
 year = {2008},
  pages = {240–243},
  doi = {10.1109/DELTA.2008.88.},
  language = {en},
  journal = {4th IEEE International Symposium on Electronic Design, Test and Applications}
}

@inproceedings{b146,
  citation-number = {144},
  author = {Sukumar, A.R. and Shah, A.F. and Anto, P.B.},
  title = {Isolated question words recognition from speech queries by using Artificial Neural Networks},
  year = {2010},
  pages = {1–4},
  language = {en},
  booktitle = {2010 Second International conference on Computing, Communication and Networking Technologies},
    doi = {10.1109/ICCCNT.2010.5591733.},
  address = {Karur, India}
}

@inproceedings{b147,
  citation-number = {145},
  author = {Rathinavelu, A. and Anupriya, G.},
  title = {Speech Recognition Model for Tamil Stops},
  pages = {4,},
 year = {2007},
  language = {en},
  booktitle={Proceedings of the World Congress on Engineering},
   Volume={1},

}

@incollection{b148,
  citation-number = {146},
  author = {Ashraf, J. and Iqbal, N. and Khattak, N.S. and Zaidi, A.M.},
  title = {Speaker Independent Urdu Speech Recognition Using HMM},
  volume = {6177},
  publisher = {Springer},
 year = {2010},
  pages = {140–148},
  doi = {10.1007/978-3-642-13881-2_14.},
  language = {en},
  booktitle = {Natural Language Processing and Information Systems},
  address = {Berlin, Heidelberg}
}

@article{b149,
  citation-number = {147},
  author = {Benisty, H. and Katz, I. and Crammer, K. and Malah, D.},
  title = {Discriminative Keyword Spotting for limited-data applications},
  volume = {99},
  pages = {1–11,},
 year = {2018},
  doi = {10.1016/j.specom.2018.02.003.},
  language = {en},
  journal = {Speech Commun}
}

@inproceedings{b152,
  citation-number = {150},
  author = {Barkani, F. and Satori, H. and Hamidi, M. and Zealouk, O. and Laaidi, N.},
  title = {Amazigh Speech Recognition Embedded System},
  language = {en},
  booktitle = {2020 1st International Conference on Innovative Research in Applied Science, Engineering and Technology, Meknes and Morocco, Apr},
 year = {2020},
  pages = {1–5},
  doi = {10.1109/IRASET48871.2020.9092014.}
}

@book{b153,
  citation-number = {151},
  author = {Downing, L.J. and Rialland, A.},
  title = {Intonation in African tone languages},
  volume = {24},
  publisher = {Walter de Gruyter GmbH and Co KG},
 year = {2016},
  language = {en}
}

@article{b154,
  citation-number = {152},
  author = {Kaur, J. and Singh, A. and Kadyan, V.},
  title = {Automatic Speech Recognition System for Tonal Languages: State-of-the-Art Survey},
  volume = {28},
 year = {2021},
  language = {en},
  journal = {Arch. Comput. Methods Eng},
  number = {3}
}

@article{b155,
  citation-number = {153},
  author = {Jia, Y.},
  title = {Direct speech-to-speech translation with a sequence-tosequence model},
 year = {2019},
  language = {en},
  journal = {ArXiv Prepr. ArXiv190406037}
}

@misc{b156,
  citation-number = {154},
  author = {Shangguan, Y. and Li, J. and Liang, Q. and Alvarez, R. and McGraw, I.},
  title = {Optimizing speech recognition for the edge},
  note = {ArXiv Prepr. ArXiv190912408,},
 year = {2019},
  language = {en}
}

@misc{b157,
  citation-number = {155},
  author = {Jongman, Suzanne R. and Khoe, Y.H. and Hintz, F.},
  title = {Vocabulary Size Influences Spontaneous Speech in Native Language Users: Validating the Use of Automatic Speech Recognition in Individual Differences Research},
   volume = {64},
  pages = {35–51},
 year = {2021},
  url = {https://doi.org/10.1177/0023830920911079.},
  doi = {10.1177/0023830920911079.},
  language = {de},
  journal = {Language and Speech},
  number = {1},
  language = {en}
}

@article{b158,
  citation-number = {156},
  author = {Yi, L. and Li, Q. and Yang, B. and Yan, Z. and Tan, H. and Chen, Z.},
  title = {Improving Speech Recognition Models with Small Samples for Air Traffic Control Systems,”},
  volume = {445},
  pages = {287–97,},
 year = {2021},
  url = {https://doi.org/10.1016/j.neucom.2020.08.092.},
  doi = {10.1016/j.neucom.2020.08.092.},
  language = {en},
  journal = {Neurocomputing},
  number = {juillet}
}

@misc{b159,
  citation-number = {157},
  author = {Jordan, G. and MacDonald, R.L. and Jiang, P. and Cattiau, J. and Heywood, R. and Cave, R. and Seaver, K. and al},
  title = { Automatic Speech Recognition of Disordered Speech: Personalized Models Outperforming Human Listeners on Short Phrases},
  year = {2021},
  note = {In Interspeech 2021, 4778-82. ISCA.},
  url = {https://doi.org/10.21437/Interspeech.2021-1384.},
  doi = {10.21437/Interspeech.2021-1384.},
  language = {en}
}

@article{b160,
  citation-number = {158},
  author = {Tejedor-García, C. and Cardeñoso-Payo, V. and Escudero-Mancebo, D.},
  title = { Automatic Speech Recognition (ASR) Systems Applied to Pronunciation Assessment of L2 Spanish for Japanese Speakers, ” Preprint},
 year = {2021},
  url = {https://doi.org/10.20944/preprints202106.0687.v1.},
  doi = {10.20944/preprints202106.0687.v1.},
  language = {en},
  journal = {PHYSICAL SCIENCES}
}

@inproceedings{b161,
  citation-number = {159},
  author = {Adnene, N. and Sabri, B. and Mohammed, B.},
  title = { Design and Implementation of an Automatic Speech Recognition Based Voice Control System},
  note = {EasyChair Preprint, Conference on Electrical Engineering 2021.},
  year = {2021},
  booktitle = {Conference on Electrical Engineering 2021},
  language = {en}

}

@article{b162,
  citation-number = {160},
  author = {Rajashri, K. and Ambewadikar, M.A. and Baheti, M.R.},
  title = { Review on small vocabulary automatic speech recognition system (asr) for marathi},
  volume = {6},
 year = {2021},
  issue = {2},
  journal={Open access International journal of Science and Engineering},
  language = {en}
}

@article{b163,
  citation-number = {161},
  author = {Mishaim, M. and Malik, M.K. and Mehmood, K. and Makhdoom, I.},
  title = { Automatic Speech Recognition: A Survey},
  volume = {80},
  pages = {9411–57,},
 year = {2021},
  doi = {10.1007/s11042-020-10073-7},
  language = {en},
  journal = {Multimedia Tools and Applications},
  number = {6}
}

@article{b164,
	title = {Improving {Speech} {Recognition} {Accuracy} for {Small} {Vocabulary} {Applications} in {Adverse} {Environments}},
	language = {en},
	author = {Thambiratnam, D and Sridharan, S},
	year = {2000},
	pages = {109–117},
	volume = {9},
	journal = {INTERNATIONAL JOURNAL OF SPEECH TECHNOLOGY},
}
\end{multicols}
\end{document}